%% file: main_arxiv.tex
\documentclass[english]{article}
\pdfoutput=1
\usepackage[T1]{fontenc}
\usepackage[utf8]{inputenc}
\usepackage{geometry}
\usepackage{float}
\usepackage{bm}
\usepackage{bbm}
\usepackage{amsmath}
\usepackage{subcaption}
\usepackage{amssymb}
\usepackage{graphicx}
\usepackage{hyperref}
\usepackage{hhline}
\hypersetup{colorlinks,linkcolor=blue,anchorcolor=brown,citecolor=blue}
\usepackage{breakurl}
\usepackage{amsthm}
\usepackage{dsfont}
\usepackage{array}
\usepackage{color}
\makeatletter
\floatstyle{ruled}
\newfloat{algorithm}{tbp}{loa}
\providecommand{\algorithmname}{Algorithm}
\floatname{algorithm}{\protect\algorithmname}

\allowdisplaybreaks
\usepackage{enumitem}
\usepackage{csquotes}
\usepackage{algorithm}
\usepackage{algorithmic}
\usepackage{arydshln}
\usepackage{cleveref}
\usepackage[misc,geometry]{ifsym} 
\numberwithin{equation}{section}
\makeatother
\DeclareFixedFont{\ttb}{T1}{txtt}{bx}{n}{10} 
\DeclareFixedFont{\ttm}{T1}{txtt}{m}{n}{10}  

\usepackage{color}
\definecolor{deepblue}{rgb}{0,0,0.5}
\definecolor{deepred}{rgb}{0.6,0,0}
\definecolor{deepgreen}{rgb}{0,0.5,0}
\definecolor{yxc}{RGB}{255,0,0}
\definecolor{yjc}{RGB}{125,0,0}
\definecolor{cm}{RGB}{0,0,200}
\definecolor{kzw}{RGB}{0,150,0}
\definecolor{codegreen}{rgb}{0,0.6,0}
\definecolor{codegray}{rgb}{0.5,0.5,0.5}
\definecolor{codepurple}{rgb}{0.58,0,0.82}
\definecolor{backcolour}{rgb}{0.95,0.95,0.92}

\newcommand{\etal}{\textit{et al}. }
\newcommand{\ie}{\textit{i}.\textit{e}., }
\newcommand{\eg}{\textit{e}.\textit{g}., }

\begin{document}

\theoremstyle{definition}

\title{An Overview of Deep Semi-Supervised Learning}

\author{%
  Yassine Ouali \Letter\thanks{Corresponding author, any corrections, contributions or suggestions are welcomed.}\hspace*{1cm}
  Céline Hudelot\hspace*{1cm} Myriam Tami\\
  Université Paris-Saclay, CentraleSupélec, MICS, 91190, Gif-sur-Yvette, France\\
  \texttt{\small \{yassine.ouali,celine.hudelot,myriam.tami\}@centralesupelec.fr}}
\date{}

\maketitle
\begin{abstract}
Deep neural networks demonstrated their ability to provide remarkable performances on a wide range of supervised learning tasks (\eg image classification) when trained on extensive collections of labeled data (\eg ImageNet). However, creating such large datasets requires a considerable amount of resources, time, and effort. Such resources may not be available in many practical cases, limiting the adoption and the application of many deep learning methods. In a search for more data-efficient deep learning methods to overcome the need for large annotated datasets, there is a rising research interest in semi-supervised learning and its applications to deep neural networks to reduce the amount of labeled data required, by either developing novel methods or adopting existing semi-supervised learning frameworks for a deep learning setting. In this paper, we provide a comprehensive overview of deep semi-supervised learning, starting with an introduction to the field, followed by a summarization of the dominant semi-supervised approaches in deep learning\footnote{A curated and an up-to-date list of SSL papers is available at this \href{https://github.com/yassouali/awesome-semi-supervised-learning}{link}.}.
\end{abstract}


\input{introduction}

\input{consistency}
\input{proxylabels}
\input{holostic}

\input{generativemodels}
\input{graphbased}

\input{unsupervised}

\section{Conclusion}
In this paper, we introduced semi-supervised learning, with its main approaches and assumptions, with SSL techniques within deep learning framework. Specifically, this review covered four broad categories of approaches for SSL: consistency regularization, generative models, graph-based methods, and holistic approaches.
With the growing research interest in data-efficient deep learning algorithms, it is foreseeable that deep SSL methods could approach the performance of fully supervised methods, and have board applications integrated into different systems and learning paradigms.

\subsection*{Acknowledgements}
Y. Ouali is supported by Randstad corporate research in collaboration with Université Paris-Saclay, CentraleSupélec, MICS. We thank Victor Bouvier for his helpful feedback on an earlier version.

\footnotesize
\bibliographystyle{plain}
\bibliography{bibSSL}

\end{document}

%% file: introduction.tex
\section{Introduction}
In recent years, semi-supervised learning (SSL) has emerged as an exciting new research direction in deep learning. Such methods deal with the situation where
few labeled training examples are available together with a significant number of unlabeled samples. In such a setting, SSL methods are more applicable to real-world applications where the unlabeled data are readily available and easy to acquire, while labeled instances are often hard, expensive, and time-consuming to collect.
SSL is capable of building better classifiers that compensate for the lack of labeled training data. However, in order to avoid a lousy matching of the problem structure with the model assumption,
which can lead to a degradation in classification performance \cite{zhu2005semi}, SSL is only effective under certain assumptions, such as assuming that the decision boundary should avoid regions with high density, facilitating the extraction of additional information from the unlabeled instances to regularize training. 
In this paper, we will start by an introduction to SSL with its main assumptions and methods, followed by a summarization of the dominant semi-supervised approaches in deep learning.
For a detailed and comprehensive review of the field, Semi-Supervised Learning Book \cite{chapelle2009semi} is a good resource.

\subsection{Semi-supervised learning}

\blockquote{
\textit{``Semi-supervised learning (SSL) is halfway between supervised and unsupervised learning.
In addition to unlabeled data, the algorithm is provided with some supervision
information – but not necessarily for all examples. Often, this information will
be the targets associated with some of the examples. In this case, the data set $X=(x_i);\ i \in [n]$
can be divided into two parts: the points $X_{l}:=(x_{1}, \dots, x_{l})$, for which labels
$Y_{l}:=(y_{1}, \dots, y_{l})$ are provided, and the points
$X_{u}:=(x_{l+1}, \ldots, x_{l+u})$, the labels of which are not known.''} -- Chapelle \etal \cite{chapelle2009semi}.}

As stated in the definition above, in SSL, we are provided with a dataset containing both
labeled and unlabeled examples. The portion of labeled examples is usually quite small
compared to the unlabeled example (\eg 1 to 10\% of the total number of examples). So with a
dataset $\mathcal{D}$ containing a labeled subset $\mathcal{D}_l$ and an unlabeled subset
$\mathcal{D}_u$, the objective, or rather hope, is to leverage the unlabeled
examples to train a better performing model than what can be obtained using only the
labeled portion. And hopefully, get closer to the desired optimal performance, in which
all of the dataset $\mathcal{D}$ is labeled.

\begin{figure}
\centering
\includegraphics[width = 0.9\textwidth]{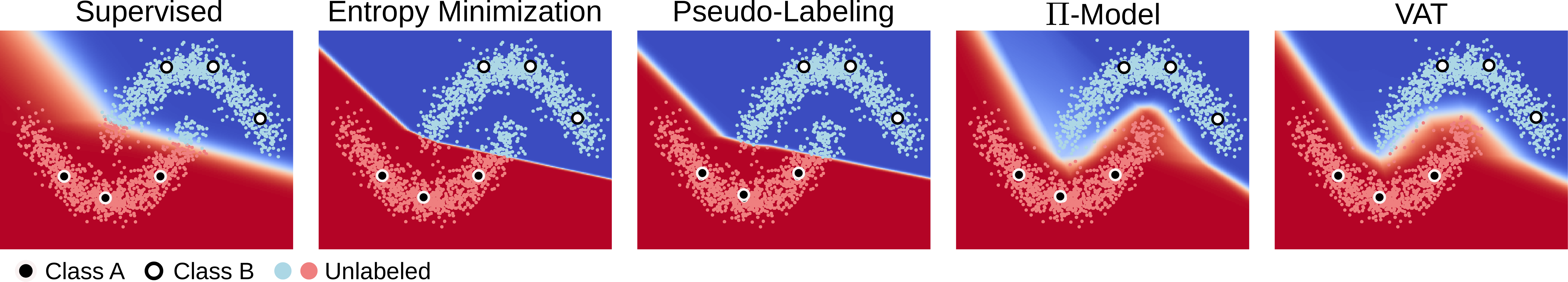}
\caption{\textbf{SSL toy example.} The decision boundaries obtained on two-moons dataset, with a supervised and different SSL approaches
using 6 labeled examples, 3 for each class, and the rest of the points as unlabeled data.}
\label{fig:ssl}
\end{figure}

More formally, the goal of SSL is to
leverage the unlabeled data $\mathcal{D}_u$ to produce a prediction function
$f_{\theta}$ with trainable parameters $\theta$, that is more accurate than what would have been obtained
by only using the labeled data $\mathcal{D}_l$.
For instance, $\mathcal{D}_u$ might provide us with additional information about the structure of the data
distribution $p(x)$ to better estimate the decision boundary
between the different classes. For example, as shown in \cref{fig:ssl}, where the data points with distinct labels are separated
with a low-density region, leveraging unlabeled data with a SSL approach can provide us with additional information
about the shape of the decision boundary between two classes, and reduce the ambiguity present in the supervised case.

SSL first appeared in the form of self-training \cite{chapelle2009semi}, which is also known as self-labeling or self-teaching.
A model is first trained on labeled data. Then, iteratively, a portion of the unlabeled data is annotated
using the trained model and added to the training set for the next training iteration. SSL took off in the 1970s after its success
with iterative algorithms such as the expectation-maximization algorithm \cite{moon1996expectation},
in which the labeled and unlabeled data are jointly used to maximize the likelihood of the model. 

\subsection{SSL Methods}

There have been many SSL methods and approaches that have been introduced over the years. These algorithms can be broadly divided into the following categories:

\begin{itemize}
\item \textbf{Consistency Regularization (a.k.a Consistency Training).} Based on the assumption that if a realistic perturbation was applied to the unlabeled data points, the prediction should not change significantly. The model can then be trained to have a consistent prediction on a given unlabeled example and its perturbed version.
\item \textbf{Proxy-label Methods.} Such methods leverage a trained model on the labeled set to produce additional training examples by labeling instances of the unlabeled set based on some heuristics. These approaches can also be referred to as \textit{bootstrapping} \cite{biemann2007unsupervised} algorithms.
We follow Ruder \etal \cite{ruder2018strong} and refer to them as proxy-label methods.
Some examples of such methods are \textit{Self-training}, \textit{Co-training} and \textit{Multi-View Learning}.
\item \textbf{Generative Models.} Similar to the supervised setting, where the learned features on one task can be transferred to other downstream tasks. Generative models that are able to generate images from the data distribution $p(x)$ must learn transferable features to a supervised task $p(y | x)$ for a given task with targets $y$.
\item \textbf{Graph-Based Methods.} The labeled and unlabeled data points can be considered as nodes of a graph, and the objective is to propagate the labels from the labeled nodes to the unlabeled ones by utilizing the similarity of two nodes $x_i$ and $x_j$, which is reflected by how strong the edge $e_{ij}$ between the two nodes.
\end{itemize}

In addition to these main categories, there is also some SSL work on \textit{entropy minimization}, where we force the model to make confident predictions by minimizing the entropy of the predictions.
Consistency training can also be considered a proxy-label method, with a subtle difference, instead of considering the predictions as ground-truths and compute the cross-entropy loss, we enforce consistency of predictions by minimizing a given distance between the outputs.

SSL methods can also be categorized based on two dominant learning paradigms, \textbf{transductive learning} and \textbf{inductive learning}. Transductive learning aims to apply the trained classifier on the unlabeled instances observed at training time; in this case, it does not generalize to unobserved instances. This type of algorithm is mainly used on graphs, such as random walks for node embedding \cite{perozzi2014deepwalk,grover2016node2vec}, where the objective is to label the unlabeled nodes of the graph that are present at training time. The more popular paradigm, inductive learning, aims to learn a classifier capable of generalizing to unobserved instances at test time.

\subsection{Main Assumptions in SSL}

The first question we need to answer is under what assumptions can we apply SSL algorithms? SSL algorithms only work under some assumptions about the structure of the data need to hold. Without such assumptions, it would not be possible to generalize from a finite training set to a set of possibly infinitely many unseen test cases. The main assumptions in SSL are:

\begin{itemize}
\item \textbf{The Smoothness Assumption.} \textit{If two points $x_1$, $x_2$ reside in a high-density region are close, then so should be their corresponding outputs $y_1$, $y_2$} \cite{chapelle2009semi}. Meaning that if two inputs are of the same class and belong to the same cluster, which is a high-density region of the input space, then their corresponding outputs need to be close. The inverse also holds true; if the two points are separated by a low-density region, the outputs must be distant from each other. This assumption can be quite helpful in a classification task, but not so much for regression.
\item \textbf{The Cluster Assumption.} \textit{If points are in the same cluster, they are likely to be of the same class} \cite{chapelle2009semi}. In this particular case of the smoothness assumption, we suppose that input data points form clusters, and each cluster corresponds to one of the output classes.
The cluster assumption can also be seen as the low-density separation assumption: \textit{The decision boundary should lie in the low-density regions.}
The relation between the two assumptions is easy to see, if a given decision boundary lies in a high-density region, it will likely cut a cluster
into two different classes, resulting in samples from different classes belonging to the same cluster, which is a violation of the cluster assumption.
In this case, we can restrict our model to have consistent predictions on the unlabeled data over some small perturbations pushing its decision boundary to low-density regions.
\item \textbf{The Manifold Assumption.} \textit{The (high-dimensional) data lie (roughly) on a low-dimensional manifold} \cite{chapelle2009semi}. In high dimensional spaces, where the volume grows exponentially with the number of dimensions, it can be quite hard to estimate the true data distribution for generative tasks. For discriminative tasks, the distances are similar regardless of the class type, making classification quite challenging. However, if our input data lies on some lower-dimensional manifold, we can try to find a low dimensional representation using the unlabeled data and then use the labeled data to solve the simplified task.
\end{itemize}

\subsection{Related Problems}

\paragraph{Active Learning}
In active learning  \cite{settles2009active,hanneke2009theoretical}, the learning algorithm is provided with a large pool of unlabeled data points, with the ability to request the labeling of any given examples from the unlabeled set in an interactive manner. As opposed to classical passive learning, in which the examples to be labeled are chosen randomly from the unlabeled pool, active learning aims to carefully choose the examples to be labeled to achieve a higher accuracy while using as few requests as possible, thereby minimizing the cost of obtaining labeled data. This is of particular interest in problems where data may be abundant, but labels are scarce or expensive to obtain.

Although it is not possible to obtain a universally good active learning strategy \cite{dasgupta2005analysis}, there exist many heuristics \cite{settles2009active}, which have been proven to be effective in practice. The two widely used selection criteria are \textit{informativeness} and \textit{representativeness} \cite{huang2010active,zhou2018brief}. \textit{Informativeness} measures how well an unlabeled instance helps reduce the uncertainty of a statistical model, while \textit{representativeness} measures how well an instance helps represent the structure of input patterns.

Active learning and SSL are naturally related, since both aim to use a limited amount of data to improve a learner. Several works considered combining SSL and AL in different tasks. \cite{drugman2019active} demonstrates a significant error reduction with limited labeled data for speech understanding, \cite{rhee2017active} proposes an active semi-supervised learning system for pedestrian detection, \cite{zhu2003combining} combines AL and SSL using Gaussian fields applied to synthetic datasets, and \cite{gao2019consistency} exploits both labeled and unlabeled data using SSL to distill information from unlabeled data that improves representation learning and sample selection.

\paragraph{Transfer Learning and Domain Adaptation}

Transfer learning \cite{pan2009survey,weiss2016survey} is used to improve a learner on one domain, called the target domain, by transferring the knowledge learned from a related domain, referred to as the source domain. For instance, we may wish to train the model on a synthetic, cheap-to-generate data, with the goal of using it on real data. 
In this case, the source domain used to train the model is related but different from the target domain used to test the model.
When the source and target differ but are related, then transfer learning can be applied to obtain higher accuracy on the target data.

One popular type of transfer learning is domain adaptation \cite{quionero2009dataset,patel2015visual,wilson2018survey}.
Domain adaptation is a type of transductive transfer learning, where the target task remains the same as the source, but the domain differs. The objective of domain adaptation
is to train a learner capable of generalizing across different domains of different distributions in which the labeled data are available for the source domain. As for the target domain, we refer to the case where no labeled data is available on target as unsupervised domain adaptation, while semi-supervised and supervised domain adaptation refers to situations where we have a limited or a fully labeled target domain receptively \cite{beijbom2012domain}.

SSL and unsupervised domain adaptation are closely related; in both cases, we are provided with labeled and unlabeled data, with the objective of learning a function capable of generalizing to the unlabeled data and unseen examples. However, in SSL, both the labeled and unlabeled sets come from the same distribution, while in unsupervised domain adaptation,
the target and source distributions differ. Methods in both subjects can be leveraged interchangeably. In SSL, \cite{mayer2019adversarial} proposed to use adversarial distribution alignment \cite{ganin2014unsupervised} for semi-supervised image classification using only a small amount of labeled samples. As for unsupervised domain adaptation, semi-supervised methods, such
as consistency regularization \cite{shu2018dirt,lee2019drop,french2017self}, co-regularization \cite{kumar2018co} or proxy labeling \cite{saito2017asymmetric,ruder2018strong} demonstrated their effectiveness in domain adaptation.

\paragraph{Weakly-Supervised Learning} 
To overcome the need for large hand-labeled and expensive training sets, most sizeable deep learning systems use some form of weak supervision: lower-quality, but larger-scale training sets constructed via strategies such as using cheap annotators \cite{ratnerweak}.
In weakly-supervised learning, the objective is the same as in supervised learning, however, instead of a ground-truth labeled training set, we are provided with
one or more weakly annotated examples, that could come from crowd workers, be the output of heuristic rules, the result of distant supervision \cite{mintz2009distant}, or the output
of other classifiers. For example, in weakly-supervised semantic segmentation, pixel-level labels, which are harder and more expensive to acquire, are substituted for inexact annotations, \eg image labels \cite{fickle,CAM,dilated_cam,wheretolook,ficklenet}, points \cite{bearman2016s}, scribbles \cite{scribblesup} and bounding boxes \cite{song2019boxdriven,Dai_2015}. In such a scenario, SSL approaches can be used to enhance the performance further if a limited number of strongly labeled examples are available while still taking advantage of the weakly labeled examples. 

\paragraph{Learning with Noisy Labels}
Learning from noisy labels \cite{frenay2013classification,garcia2015data} can be challenging given
the negative impact label noise can have on the performance of deep learning methods if the noise is significant.
To overcome this, most existing methods for training deep neural networks with noisy labels seek to correct the loss function.
One type of correction consists of treating all the examples as equal and relabeling the noisy examples, where proxy labels methods
can be used for the relabeling procedure \cite{yi2019probabilistic,ma2018dimensionality,reed2014training}. Another type of correction applies a reweighing
to the training examples to distinguish between the clean and noisy samples \cite{cooke2001robust,thulasidasan2019combating}. Other works \cite{ding2018semi,hendrycks2019augmix,kong2019recycling,li2020dividemix} have shown that SSL can be useful in learning from noisy labels, where the noisy labels are discarded, and the noisy examples are considered as unlabeled data and used to regularize training using SSL methods.

\subsection{Evaluating SSL Approaches}
The conventional experimental procedure used to evaluate SSL methods consists of choosing a dataset (\eg CIFAR-10 \cite{krizhevsky2009learning}, SVHN \cite{netzer2011reading}, ImageNet \cite{deng2009imagenet}, IMDb \cite{maas2011learning}, Yelp review \cite{zhang2015character}) commonly used for supervised learning, a large portion of the labels 
are then ignored, resulting in a small labeled set $\mathcal{D}_l$ and a larger unlabeled $\mathcal{D}_u$. A deep learning model is trained with a given SSL approach, and the results are reported on the original test set over various and standardized portions of labeled examples. In order to make this procedure applicable to real-world settings,
Oliver \etal \cite{oliver2018realistic} proposed the following ways to improve this experimental methodology:
\begin{itemize}
\item \textbf{A Shared Implementation.} For a realistic comparison of different SSL methods, they must share the same underlying architectures and
other implementation details (\eg hyperparameters, parameter initialization, data augmentation, regularization, etc.).
\item \textbf{High-Quality Supervised Baseline.} The main objective of SSL is to obtain better performance than what can be obtained in a supervised manner. This is why it is essential to provide a strong baseline consisting of training the same model on the labeled set $\mathcal{D}_l$ in a supervised way, with modified hyperparameters to report the best-case performance of the fully-supervised model.
\item \textbf{Comparison to Transfer Learning.} Another robust baseline to compare SSL methods to can be obtained by training the model on large labeled datasets, and then fine-tune it on the small labeled set $\mathcal{D}_l$.
\item \textbf{Considering Class Distribution Mismatch.} The possible distribution mismatch between the labeled and unlabeled examples can be ignored when doing evaluation since both sets come from the same dataset. Still, such a mismatch is prevalent in real-world applications, where the unlabeled data can have different class distributions compared to the labeled data. The effect of this discrepancy needs to be addressed for better real-world adoption of SSL.
\item \textbf{Varying the Amount of Labeled and Unlabeled Data.} A common practice in SSL is varying the number of labeled examples, but also varying the size $\mathcal{D}_u$ in a systematic way to simulate realistic scenarios, such as training on a relatively small unlabeled set, can provide additional insights into the effectiveness of SSL approaches.
\item \textbf{Realistically Small Validation Sets.} In many cases where a fully annotated dataset if used for evaluation, we might end-up with a validation set that is significantly larger than the labeled set $\mathcal{D}_l$ used for training, in such a setting, extensive hyperparameter tuning might result in an overfitting to the validation set. In contrast, small validation sets constrain the ability to select models \cite{chapelle2009semi,forster2018neural}, resulting in a more realistic assessment of the performance of SSL methods.
\end{itemize}

%% file: consistency.tex
\section{Consistency Regularization}

A recent line of works in deep semi-supervised learning utilizes the unlabeled data
to enforce the trained model to be in line with the cluster assumption, \ie, the
learned decision boundary must lie in low-density regions. These methods are based
on a simple concept that, if a realistic perturbation was to be applied to an unlabeled
example, the prediction should not change significantly, given that under the
cluster assumption, data points with distinct labels are separated with low-density regions,
so the likelihood of one example to switch classes after a perturbation is small (\eg \cref{fig:ssl}).

More formally, with consistency regularization, we are favoring functions $f_\theta$ that give consistent
predictions for similar data points. So rather than minimizing the classification cost at the zero-dimensional data points
of the inputs space, the regularized model minimizes the cost on a manifold around each data point, pushing the
decision boundaries away from the unlabeled data points and smoothing the manifold on which the data
resides \cite{zhu2005semi}. Concretely,
given an unlabeled data point $x \in \mathcal{D}_u$ and its perturbed version $\hat{x}_u$,
the objective is to minimize the distance between the two outputs
$d(f_{\theta}(x), f_{\theta}(\hat{x}))$. The popular distance measures $d$ are
mean squared error (MSE), Kullback-Leiber divergence (KL)
and Jensen-Shannon divergence (JS). For two
outputs $f_{\theta}(x)$ and $f_{\theta}(\hat{x})$ in the form of a probability distribution over the $C$
classes,
and $m=\frac{1}{2}(f_{\theta}(x) + f_{\theta}(\hat{x}))$, we can compute these measures as follows:
\begin{equation}
d_{\mathrm{MSE}}(f_{\theta}(x), f_{\theta}(\hat{x}))=\frac{1}{C} \sum_{k=1}^{C}(f_{\theta}(x)_k -f_{\theta}(\hat{x})_k)^{2}
\end{equation}
\begin{equation}
d_{\mathrm{KL}}(f_{\theta}(x), f_{\theta}(\hat{x}))=
\frac{1}{C} \sum_{k=1}^{C} f_{\theta}(x)_k \log \frac{f_{\theta}(x)_k}{f_{\theta}(\hat{x})_k}
\end{equation}
\begin{equation}
d_{\mathrm{JS}}(f_{\theta}(x), f_{\theta}(\hat{x}))=\frac{1}{2}
d_{\mathrm{KL}}(f_{\theta}(x), m)+\frac{1}{2} d_{\mathrm{KL}}(f_{\theta}(\hat{x}), m)
\end{equation}

Note that we can also enforce a consistency over two perturbed versions of $x$,
$\hat{x}_1$ and $\hat{x}_2$.

\subsection{Ladder Networks}

\begin{figure}
\centering
\includegraphics[width = 0.75\textwidth]{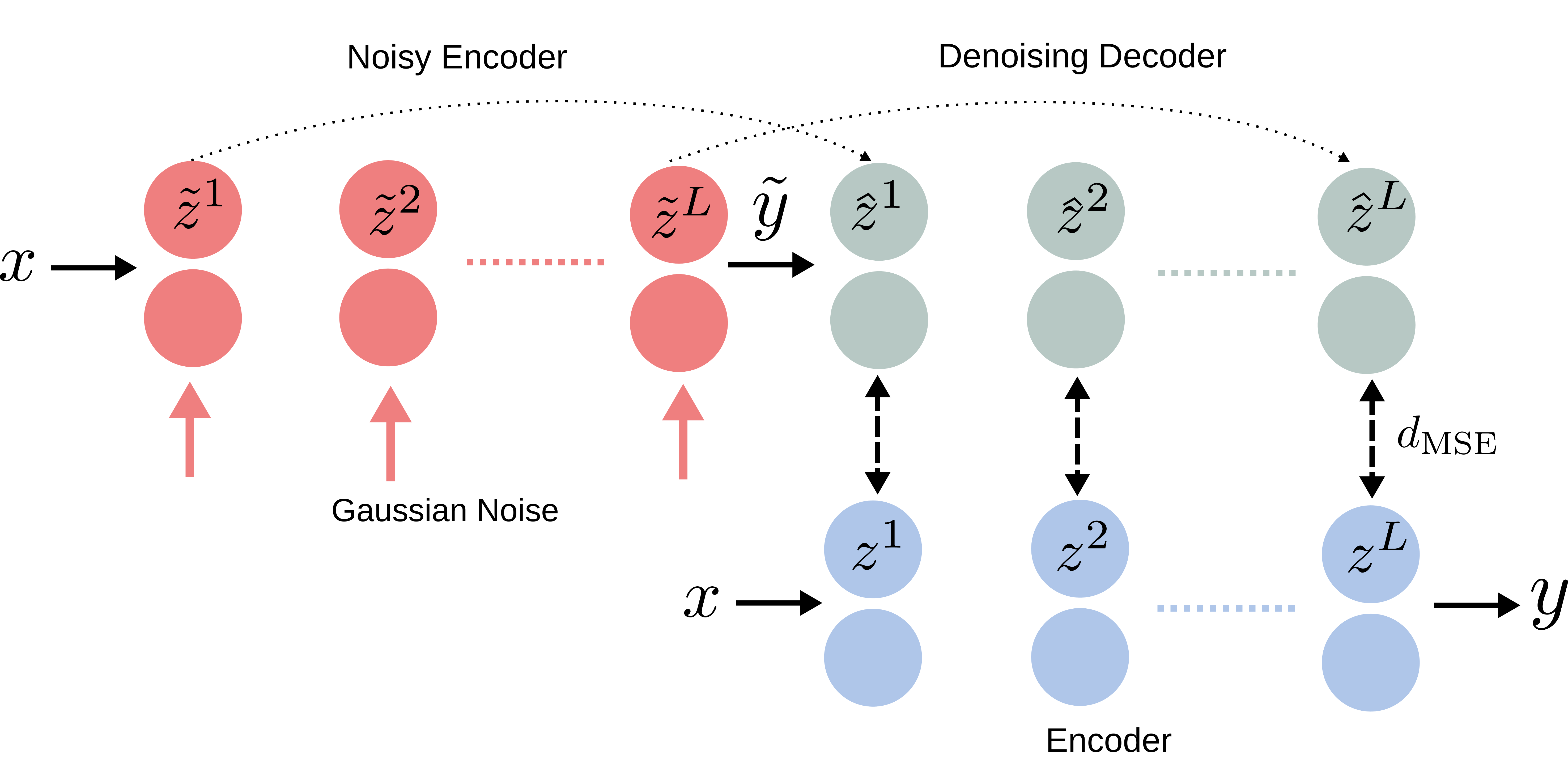}
\caption{\textbf{Ladder Networks.} An illustration of one forward pass of Ladder Networks. The objective is to reconstruct the clean activations of the encoder
using a denoising decoder that takes as input the corrupted activations of the noisy encoder.}
\label{fig:lader}
\end{figure}

To take any well-performing feed-forward network on supervised data and augment it with
additional branches to be able to utilize additional unlabeled data. Rasmus \etal \cite{rasmus2015semi} propose to use Ladder
Networks \cite{valpola2015neural} with an additional encoder and decoder for SSL.
As illustrated in \cref{fig:lader}, the network consists of two encoders, a corrupted and clean one, and a decoder.
At each training iteration, the input $x$ is passed through both encoders. In the corrupted encoder, 
Gaussian noise is injected at each layer after batch normalization, producing two outputs, a clean prediction
$y$ and a prediction based on corrupted activations $\tilde{y}$. The output $\tilde{y}$ is then fed into
the decoder to reconstruct the uncorrupted input and the clean hidden activations.
The unsupervised training loss $\mathcal{L}_u$
is then computed as the MSE between the activations of the clean encoder $z$
and the reconstructed activations $\hat{z}$ (\ie after batch normalization), computed over all layers,
from the input to the last layer $L$, with a weighting $\lambda_{l}$ for each layer's contribution to the total loss:
\begin{equation}
\mathcal{L}_u = \frac{1}{|\mathcal{D}|} \sum_{x \in \mathcal{D}}
\sum_{l=0}^{L} \lambda_{l} d_{\mathrm{MSE}}(z^{(l)}, \hat{z}^{(l)})
\end{equation}

If the input is a labeled data point, $x \in \mathcal{D}_l$, with a label $y$, a supervised cross-entropy loss $\mathrm{H}(\tilde{y}, t)$ term can be added to $\mathcal{L}_u$ to obtain the final loss.
\begin{equation}
\mathcal{L} = \mathcal{L}_u  + \mathcal{L}_s = \mathcal{L}_u +
\frac{1}{|\mathcal{D}_l|} \sum_{x, t \in \mathcal{D}_l} \mathrm{H}(\tilde{y}, t)
\end{equation}

The method can be easily adapted for convolutional neural networks (CNNs)
by replacing the fully-connected layers with
convolutional layers for semi-supervised vision tasks.
However, the ladder network is quite computationally heavy, approximately tripling
the computation needed for one training iteration. To mitigate this,
the authors propose a variant of ladder networks called \textbf{$\Gamma$-Model} where
$\lambda_{l}=0$ when $l<L$. In this case, the decoder is omitted, and the unsupervised loss
is computed as the MSE between the two outputs $y$ and $\tilde{y}$.

\subsection{Pi-Model}

\begin{figure}
\centering
\includegraphics[width = 0.8\textwidth]{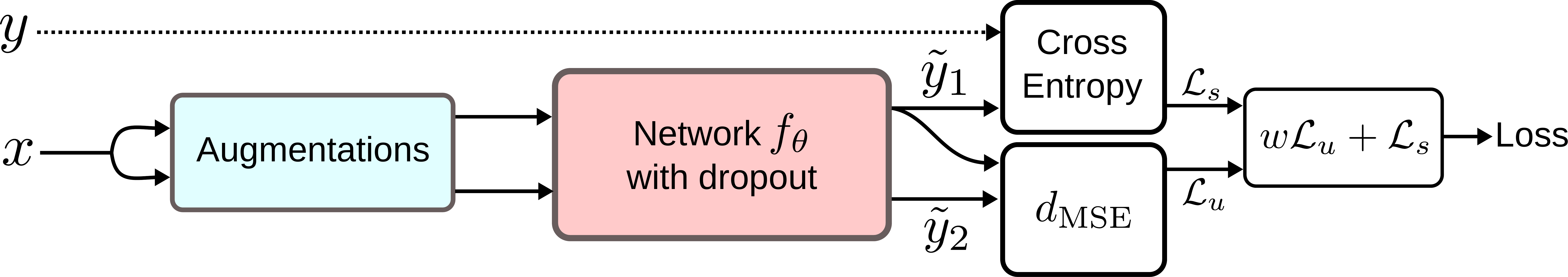}
\caption{\textbf{Loss computation for $\Pi$-Model.} The MSE between the two outputs is computed for the unsupervised loss, and if the input is a labeled example, we add the supervised loss to the weighted unsupervised loss.}
\label{fig:pimodel}
\end{figure}

The \textbf{$\Pi$-Model} \cite{laine2016temporal} is a simplification of the $\Gamma$-Model of Ladder Networks,
where the corrupted encoder is removed, and the same network is used to get the prediction for both corrupted and uncorrupted inputs.
Specifically, $\Pi$-Model takes advantage of the stochastic nature of the prediction function $f_ \theta$ in
neural networks due to common regularization techniques, such as data augmentation and dropout, that typically don't alter the model predictions. For any given input $x$, the objective is to reduce the distances between two predictions of $f_ \theta$ with $x$ as input in both forward passes. Concretely, as illustrated in \cref{fig:pimodel}, we would like to minimize $d(y, \tilde{y})$, where we consider one of the two outputs as a target. Given the stochastic nature of the predictions function (\eg using dropout as a noise source),
the two outputs $f_\theta(x) = \tilde{y}_1$ and $f_\theta(x) = \tilde{y}_2$ will be distinct, and the objective is
to obtain consistent predictions for both of them. In case the input $x$ is a labeled data point,
we also compute the cross-entropy supervised loss using the provided labels $y$:
\begin{equation}
\mathcal{L} = w \frac{1}{|\mathcal{D}_u|} \sum_{x \in \mathcal{D}_u}
d_{\mathrm{MSE}}(\tilde{y}_1, \tilde{y}_2) + 
\frac{1}{|\mathcal{D}_l|} \sum_{x, y \in \mathcal{D}_l} \mathrm{H}(y, f(x))
\end{equation}
with $w$ as a weighting function, starting from 0 up to a fixed weight $\lambda$ (\eg 30) after a
given number of epochs (\eg 20\% of training time). This way, we avoid using the untrained and
random prediction function, providing us with unstable predictions at the start of training.

\subsection{Temporal Ensembling}

$\Pi$-Model can be divided into two stages, we first classify all of the training data without updating the weights of the model,
obtaining the predictions $y$, and in the second stage, we consider the predictions $y$ as targets for the unsupervised
loss and enforce consistency of predictions by minimizing the distance between the current outputs $\tilde{y}$ and the outputs of
the first stage $y$ under different dropouts and augmentations.

\begin{figure}[htb]
\centering
\includegraphics[width = 0.8\textwidth]{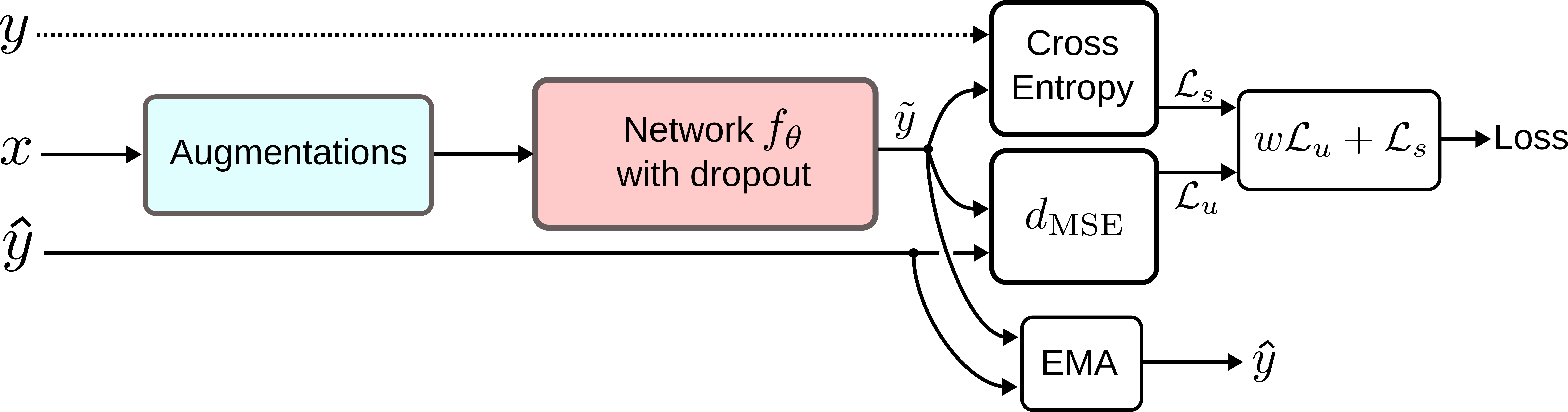}
\caption{\textbf{Loss computation for Temporal Ensembling.} The MSE between the current prediction and 
the aggregated target is computed for the unsupervised loss, and if the input is a labeled example, we add the supervised loss to the weighted unsupervised loss.}
\label{fig:temporalensembling}
\end{figure}

The problem with this approach is that the targets $y$ are based on a single evaluation of the network and can
rapidly change. This instability in the targets can lead to an instability during training and reduces the amount of training signal that can be extracted from the unlabeled examples. To solve this, Laine \etal \cite{laine2016temporal} propose a second
version of $\Pi$-Model called \textbf{Temporal Ensembling}, where the targets $y_{\mathrm{ema}}$ are the aggregation of all the previous predictions. This way, during training, we only need a single forward pass to get the current predictions $\tilde{y}$ and the aggregated targets $y_{\mathrm{ema}}$, speeding up the training time by approximately 2$\times$. The training process is illustrated in \cref{fig:temporalensembling}.

For a target $\tilde{y}$, at each training iteration, the current output $\tilde{y}$ is accumulated into the \textit{ensemble output} $y_{\mathrm{ema}}$ 
by an exponentially moving average update:
\begin{equation}
y_{\mathrm{ema}} = \alpha y_{\mathrm{ema}}+(1-\alpha) \tilde{y}
\end{equation}
where $\alpha$ is a momentum term that controls how far the ensemble reaches into training history. $\tilde{y}$ 
can also be seen as the output of an ensemble network $f$ from previous training epochs, with the recent ones having greater weight than the distant ones.

At the start of training, temporal ensembling reduces to $\Pi$-Model since the aggregated targets are very noisy,
to overcome this, similar to the bias correction used in Adam optimizer \cite{kingma2014adam}, the targets $\tilde{y}$ are corrected for the startup bias at a training step $t$ as follows:
\begin{equation}
y_{\mathrm{ema}} = (\alpha y_{\mathrm{ema}}+(1-\alpha) \tilde{y}) / (1-\alpha^{t})
\end{equation}

The loss computation in temporal ensembling remains the same as in $\Pi$-Model, but with two essential benefits. First, the training is faster since we only need a single forward pass through the network to obtain $\tilde{y}$, while maintaining
an exponential moving average (EMA) of label predictions on each training example and penalizing predictions that are inconsistent with these targets. Second, the targets are more stable during training, yielding better results. 
The downside of such a method is a large amount of memory needed to keep an aggregate of the predictions for all of the training examples,
which can become quite memory intensive for large datasets and dense tasks (\eg semantic segmentation).

\subsection{Mean teachers}

$\Pi$-Model and its improved version with Temporal Ensembling provides a better and more stable teacher model by maintaining an EMA of the predictions of each example, formed by an ensemble of the model's current version and those earlier versions evaluated at the same example. This ensembling improves the quality of the predictions and using them as the teacher predictions improve results. However, the newly learned information is incorporated into the training at a slow pace, since each target is updated only once per epoch, and the larger the dataset, the bigger the span between the updates gets.

\begin{figure}[htb]
\centering
\includegraphics[width = 0.7\textwidth]{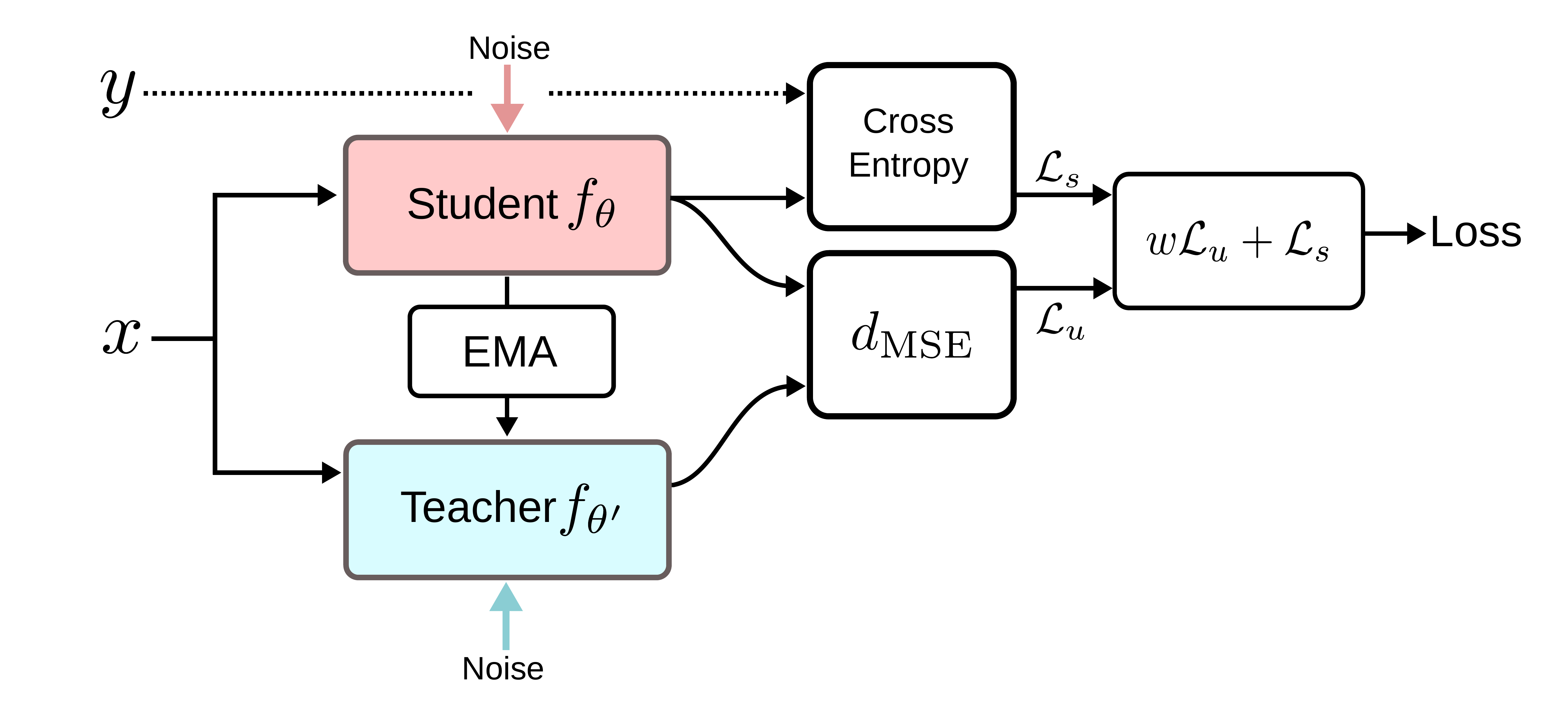}
\caption{\textbf{Mean Teacher.} The teacher model, which is an EMA of the student model, is responsible for generating the targets for consistency training. The student model is then trained to minimize the supervised loss over labeled examples and the consistency loss over unlabeled examples. At each training iteration, both
models are evaluated with an injected noise ($\eta$, $\eta^\prime$), and the weights of the teacher model are updated using the current student model to incorporate the learned information at a faster pace.}\label{fig:meanteacher}
\end{figure}

Additionally, in the previous approaches, the same model plays a dual role, as a \textit{teacher} and a \textit{student}. Given a set of 
unlabeled data, as a teacher, the model generates the targets, which are then used by itself as a student for learning using a consistency loss. These targets may very well be misclassified. If the weight of the unsupervised loss outweighs that of the supervised loss,
the model is prevented from learning new information, predicting the same targets, and resulting in a form of confirmation bias. To solve this, the quality of the targets must be improved. The quality of targets can be improved by either: (1) carefully choosing the perturbations instead of merely injecting additive or multiplicative noise, or, (2) carefully choosing the teacher model responsible for generating the targets, instead of using a replica of the student model.

To overcome these limitations, Mean Teacher \cite{tarvainen2017mean} proposes using a teacher model for a faster incorporation of the learned signal, and to avoid the problem of confirmation bias.
A training iteration of Mean Teacher (\cref{fig:meanteacher}) is very similar to previous methods; the main difference is that $\Pi$-Model uses the same model as a student and a teacher $\theta^{\prime}=\theta$, and Temporal Ensembling approximate a stable teacher $ f_{\theta^{\prime}}$ as an ensemble function with a weighted average of successive predictions. While Mean Teacher defines the weights $ \theta^{\prime}_t$  of the teacher model $f_{\theta^{\prime}}$ at a training step $t$  as an EMA of successive student's weights $ \theta$:
\begin{equation}
\theta_{t}^{\prime}=\alpha \theta_{t-1}^{\prime}+(1-\alpha) \theta_{t}
\end{equation}

The loss computation in this case is the sum of the supervised and unsupervised loss, where the teacher model is used to obtain the targets
for the unsupervised loss for a given input $x$:
\begin{equation}
\mathcal{L} = w \frac{1}{|\mathcal{D}_u|} \sum_{x \in \mathcal{D}_u}
d_{\mathrm{MSE}}(f_{\theta}(x), f_{\theta^{\prime}}(x)) + 
\frac{1}{|\mathcal{D}_l|} \sum_{x, y \in \mathcal{D}_l} \mathrm{H}(y, f_{\theta}(x)) 
\end{equation}

\subsection{Dual Students}


One of the main drawbacks of using a Mean Teacher is that given a large number of training iterations, the teacher model weights will converge to that of the student model, and any biased and unstable predictions will be carried over to the student.

To solve this, Ke \etal \cite{ke2019dual} propose a dual students step-up. Two student models with different initialization are simultaneously trained, and at a given iteration, one of them
provides the targets for the other. To choose which one, we test for the most stable predictions that satisfy
the following stability conditions:
\begin{itemize}
\item The predictions using two input versions, a clean $x$  and a perturbed version $ \tilde{x}$  give
the same results: $f(x) = f(\tilde{x})$.
\item Both predictions are confident, \ie are far from the decision boundary. This
can be tested by seeing if $f(x)$ (resp. $ f(\tilde{x})$) is greater than a confidence threshold $\epsilon$, \eg $\epsilon=0.1$.
\end{itemize}

Given two student models, $f_{\theta_1}$ and $f_{\theta_2}$, an unlabeled input $x \in \mathcal{D}_u$ and its perturbed version $\tilde{x}$. We compute four predictions: $ f_{\theta_1}(x), f_{\theta_1}(\tilde{x}), f_{\theta_2}(x)$, and $f_{\theta_2}(\tilde{x})$ .
In addition to training each model to minimize both the supervised and unsupervised losses:
\begin{equation}
\mathcal{L} = \mathcal{L}_s + \lambda_1 \mathcal{L}_u
= \frac{1}{|\mathcal{D}_l|} \sum_{x, y \in \mathcal{D}_l} \mathrm{H}(y, f_{\theta_i} (x)) + \lambda_1
\frac{1}{|\mathcal{D}_u|} \sum_{x \in \mathcal{D}_u} d_{\mathrm{MSE}} (f_{\theta_i}(x), f_{\theta_i}(\tilde{x}))
\end{equation}
we also force one of the students to have similar predictions to its counterpart. To chose which one to update its weights, 
we check for both models' stability constraint; if the predictions one of the models is unstable, we update its weights.
If both are stable, we update the model with the largest variation $\mathcal{E}^{i} = \|f_i(x)-f_i(\tilde{x}) \|^{2}$, \ie the least stable. In this case, the least stable model is trained with an additional loss:
\begin{equation}
\lambda_2 \sum_{x \in \mathcal{D}_u} d_{\mathrm{MSE}}(f_{\theta_i}(x), f_{\theta_j}(x)) 
\end{equation}
where $\lambda_1$ and $\lambda_2$ are hyperparameters specifying the contribution of each loss term.

\subsection{Fast-SWA}

Athiwaratkun \etal \cite{athiwaratkun2018there} observed that $\Pi$-Model and Mean Teacher continue taking significant steps in the weight space at the end of training, given that the models stochastic gradient descent (SGD) traverses a large flat region of the weight space late in training,
continuing to actively explore the set of plausible solutions and producing diverse predictions on the test set even in the late stages of training. Based on this behavior, averaging the SGD iterates can lead to final weights closer to the center of the flat region, stabilizing the SGD trajectory, and leading to significant gains in performance and better generalization.

\begin{figure}[htb]
\centering
\includegraphics[width=1\textwidth]{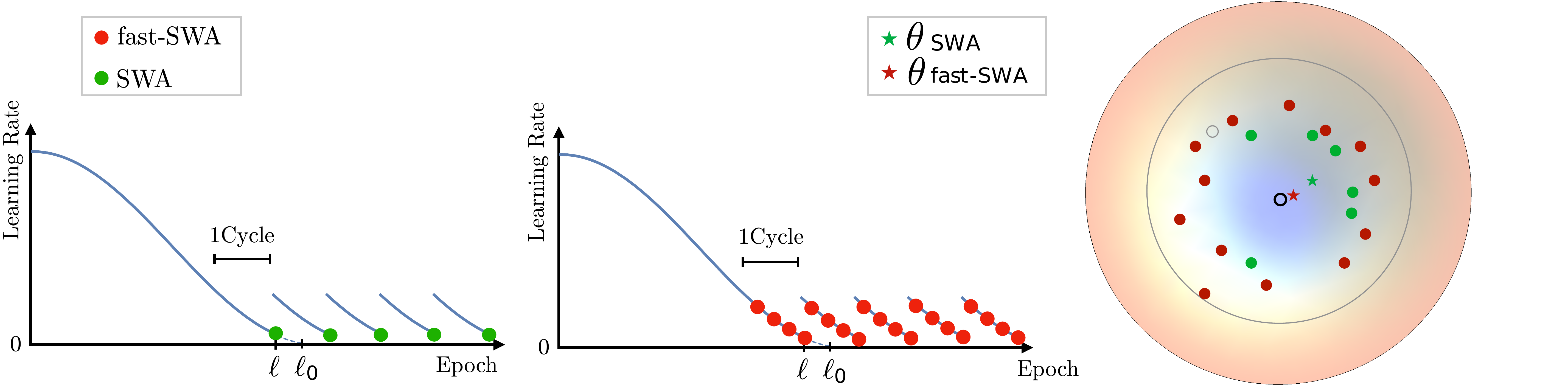}
\caption{\textbf{SWA and fast-SWA.} \textit{Left and Center.} Cyclical cosine learning rate schedule used at the end of training for SWA and fast-SWA with
different averaging strategies. \textit{Right.} 2d illustration of the impact of SWA and fast-SWA averaging strategies on the final weights.
Based on \cite{athiwaratkun2018there}.}
  \label{fig:fast-SWA}
\end{figure}

One way to produce an ensemble of the model late in training is Stochastic Weight Averaging (SWA) \cite{izmailov2018averaging}, an approach based on averaging the weights traversed by SGD at the end of training with a cyclic learning rate (\cref{fig:fast-SWA}). After a given number of epochs, the learning rate changes to a cyclic learning rate and the training repeats for several cycles, the weights at the end of each cycle corresponding to the minimum values of the learning rate are stored, and averaged together to obtain a model with the averaged weights $f_{\theta_{\text{SWA}}}$, which is then used to make predictions.

Motivated by the observation that the benefits of averaging are the most prominent when the distance between the averaged points is large, and given that SWA only collects the weights once per cycle, which means that many additional training epochs are needed in order to collect enough weights for averaging. The authors propose \textbf{fast-SWA}, a modification of SWA that averages the networks corresponding to many points during the same cycle, resulting in a better final model and a faster ensembling procedure.

\subsection{Virtual Adversarial Training}
The previous approaches focused on applying random perturbations to each input to generate artificial input points,
encouraging the model to assign similar outputs to the unlabeled data points and their perturbed versions. This way,
we push for a smoother output distribution. As a result, the generalization performance of the model can be improved. However, such 
random noise and random data augmentation often leaves the predictor particularly vulnerable to small perturbations in a specific direction, that is, the adversarial direction, which is the direction in the input space in which the label probability $p(y|x)$ of the model is most sensitive.

To overcome this, and inspired by adversarial training \cite{goodfellow2014explaining} that trains the model to assign to each input data a label that is similar to the labels of its neighbors in the adversarial direction. Miyato \etal\cite{miyato2018virtual} propose Virtual Adversarial Training (VAT), a regularization technique that enhances the model's robustness around each input data point against random and local perturbations, the term \textit{virtual} comes from the fact that the adversarial perturbation is approximated without any label information, and is hence applicable to SSL to smooth the output distribution.

Concretely, VAT trains the output distribution to be identically smooth around each data point, by selectively smoothing the model in its most adversarial direction. For a given data point $x$, we would like to compute the adversarial perturbation $r_{adv}$ that will alter the model's predictions the most. We start by sampling a Gaussian noise $r$ of the same dimensions as the input $x$, we then compute its gradients $grad_r$ with respect the loss between the two predictions, with and without the injections of the noise $r$
(\ie KL-divergence is used as a distance measure $d(.,.)$). $r_{adv}$ can then be obtained by normalizing and scaling $grad_r$ by a hyperparameter $\epsilon$. The computation can be summarized in the following steps:
\begin{enumerate}
  \item $r \sim \mathcal{N}(0, \frac{\xi}{\sqrt{\operatorname{dim}(x)}} I)$
  \item $grad_{r}=\nabla_{r} d_{\mathrm{KL}}(f_{\theta}(x), f_{\theta}(x+r))$
  \item $r_{adv}=\epsilon \frac{grad_{r}}{\|grad_{r}\|}$
\end{enumerate}

Note that the computation above is a single iteration of the approximation of $r_{adv}$, for a more accurate estimate, we consider $r_{adv} = r$ and recompute $r_{adv}$ following the last two steps. But in general, given how computationally expensive this computation is, requiring an additional forward and backward passes, we only apply a single power iteration for computing the adversarial perturbation.
With the optimal perturbation $r_{adv}$, we can then compute the unsupervised loss as the MSE between the two predictions of the model, with and without the injection of $r_{adv}$:
\begin{equation}
\mathcal{L}_u = w \frac{1}{|\mathcal{D}_u|} \sum_{x \in \mathcal{D}_u}
d_{\mathrm{MSE}}(f_{\theta}(x), f_{\theta}(x + r_{adv}))
\end{equation}

For a more stable training, a Mean Teacher can be used to generate stable targets by replacing $f_{\theta}(x)$ with $f_{\theta^{\prime}}(x)$, where $f_{\theta^{\prime}}$ is an EMA of the student $f_{\theta}$.
\begin{figure}
\centering
\includegraphics[width = 0.7\textwidth]{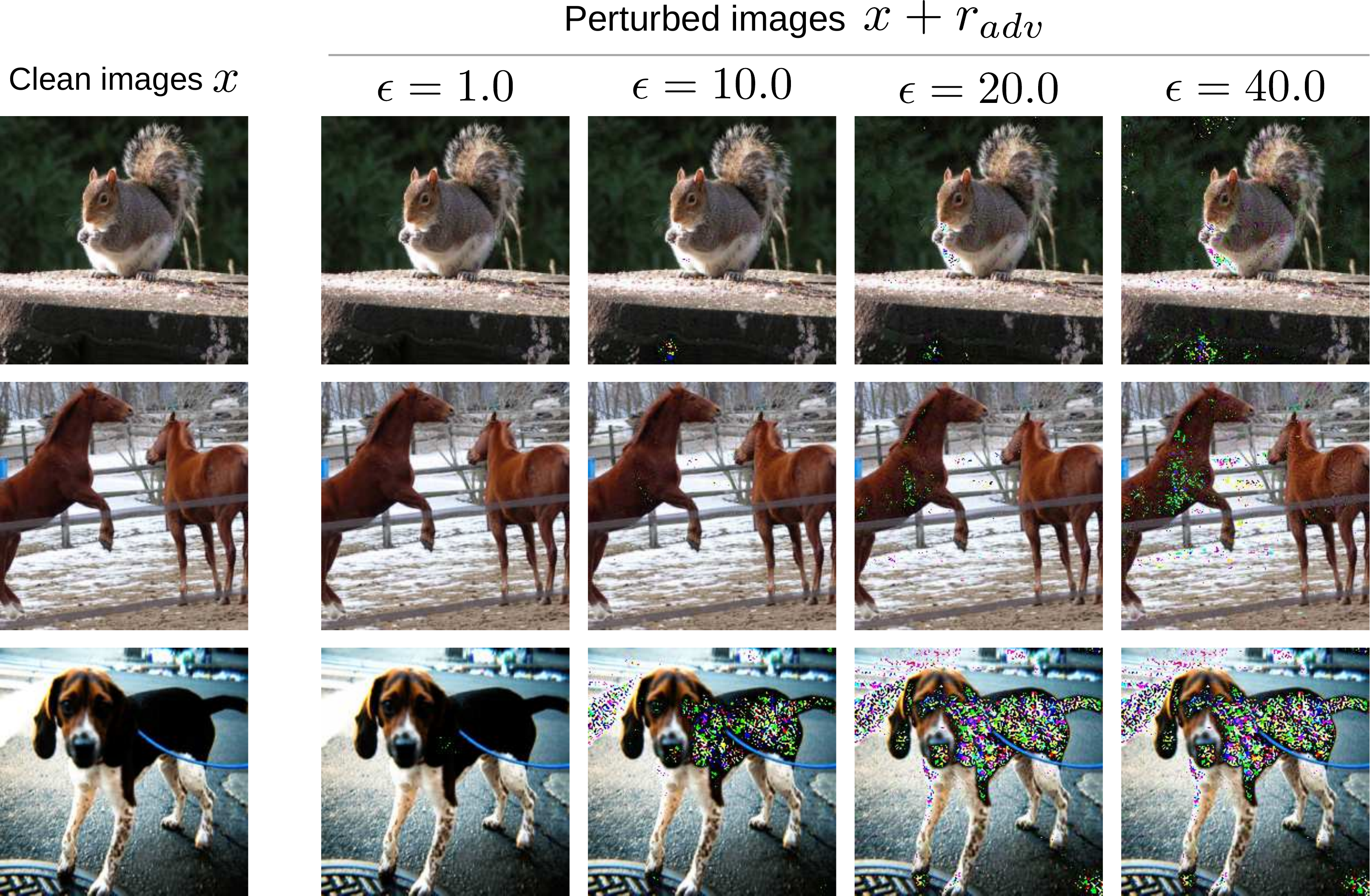}
\caption{\textbf{Virtual Adversarial Examples.} Examples of the perturbed ImagetNet images for different values of the scaling hyperparameter $\epsilon$.}
  \label{fig:vat}
\end{figure}

\subsection{Adversarial Dropout}
Instead of using an additive adversarial noise as VAT, Park \etal \cite{park2018adversarial}
propose adversarial dropout (AdD), a.k.a, element-wise adversarial dropout (EAdD), 
in which dropout masks are adversarially optimized to alter the model's predictions.
With this type of perturbations, we induce a sparse structure of the neural network, while the other forms of additive noise does not
make changes to the structure of the neural network directly.

The first step is to find the dropout conditions that are most sensitive to the model's predictions. In a SSL setting, where we do not have access to the true labels, we use the model's predictions on the unlabeled data points to approximate
the adversarial dropout mask $\epsilon^{adv}$, which is subject to the boundary condition:
$\|\epsilon^{adv}-\epsilon\|_{2} \leq \delta H$ with $H$
as the dropout layer dimension and a hyperparameter $\delta$,
which restricts adversarial dropout masks to be infinitesimally different from the random dropout mask $\epsilon$.
Without this constraint, the adversarial dropout might induce a layer without any connections.
By restricting the adversarial dropout to be similar to the random dropout,
we prevent finding such an irrational layer, which does not support backpropagation. 

Similar to VAT, we start from a random dropout mask, we compute a KL-divergence loss
between the outputs, with and without dropout, and given the gradients of the loss with respect to the activations
before the dropout layer, we update the random dropout mask in an adversarial manner. 
The prediction function $f_{\theta}$ is divided into two parts, $f_{\theta_1}$
and $f_{\theta_2}$, where $f_{\theta}(x, \epsilon)=f_{\theta_{2}}(f_{\theta_{1}}(x) \odot \epsilon)$, 
we start by computing an approximation of the Jacobian matrix as follows:
\begin{equation}
J(x, \epsilon) \approx f_{\theta_{1}}(x)\odot
\nabla_{f_{\theta_{1}}(x)} d_{\mathrm{KL}}(f_{\theta}(x),
f_{\theta}(x, \epsilon))
\end{equation}

Using $J(x, \epsilon)$, we can then update the random dropout mask $\epsilon$
to obtain $\epsilon^{adv}$, so that if $\epsilon(i) = 0$ and $J(x, \epsilon)(i) > 0$
or $\epsilon(i) = 1$ and $J(x, \epsilon)(i) < 0$ at a given position $i$, we inverse the
value of $\epsilon$ at that location. Resulting in $\epsilon^{adv}$, which can then
be used to compute the unsupervised loss:
\begin{equation}
\mathcal{L}_u = w \frac{1}{|\mathcal{D}_u|} \sum_{x \in \mathcal{D}_u}
d_{\mathrm{MSE}}(f_{\theta}(x), f_{\theta}(x, \epsilon^{adv}))
\end{equation}

\begin{figure}[htb]
\centering
\includegraphics[width = 0.9\textwidth]{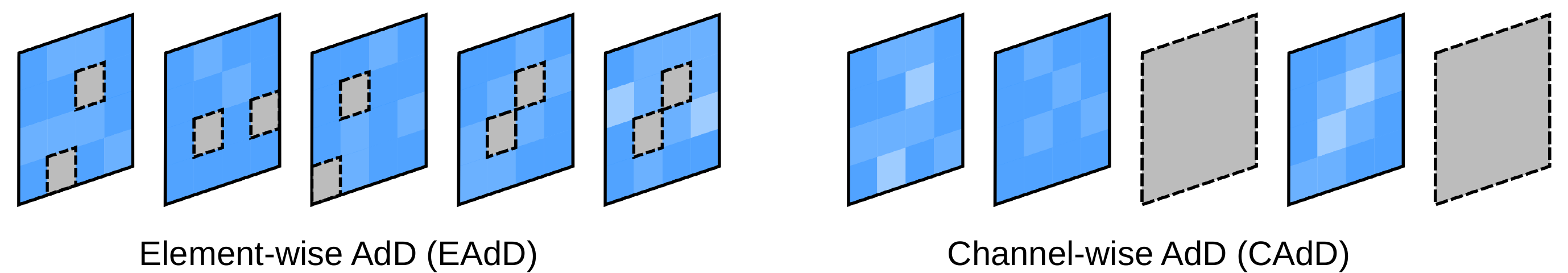}
\caption{\textbf{EAdD and CAdD.} EAdD drops activation individually regardless of the spatial correlation, while CAdD drops entire feature maps, making it more suitable for convolutional layers. Image Source: \cite{lee2019drop}.}
  \label{fig:AdD}
\end{figure}

\paragraph{Channel-wise Adversarial Dropout} The element-wise adversarial dropout (EAdD) introduced by Park  \etal \cite{park2018adversarial} is limited
to fully-connected networks, to use AdD in a wider range of tasks, Lee \etal \cite{lee2019drop} proposed channel-wise AdD (CAdD), an extension the element-wise masking in AdD to convolutional layers (\cref{fig:AdD}). In these layers, standard dropout is relatively ineffective due to the strong spatial correlation between individual activations of a feature map \cite{tompson2015efficient}. EAdD dropout suffers from the same issues when naively applied to convolutional layers. To solve this, EAdD adversarially drops entire feature maps rather than individual activations. While the general procedure is similar to that of EAdD, an additional constraint is imposed on the mask to represent spatial dropout \cite{tompson2015efficient}. In this case, the mask $\epsilon \in \mathbb{R}^{C \times H \times W}$ is of the same shape as the activations; the adversarial dropout mask $\epsilon^{adv}$ is approximated under the following new condition:
\begin{equation}
\frac{1}{H W} \sum_{i=1}^{C}\|\epsilon^{adv}(i)- \epsilon(i)\| \leq \delta C
\end{equation}
where $\delta$ is a hyperparameter to restrict the different between the two masks to be small, and $\epsilon(i)$ is the mask corresponding to the $i$-th activation map. The process of finding the channel-wise adversarial dropout mask is similar to those of element-wise adversarial dropout, but with a per activation map approximation.

\subsection{Interpolation Consistency Training}

\begin{figure}
\centering
\includegraphics[width = 0.8\textwidth]{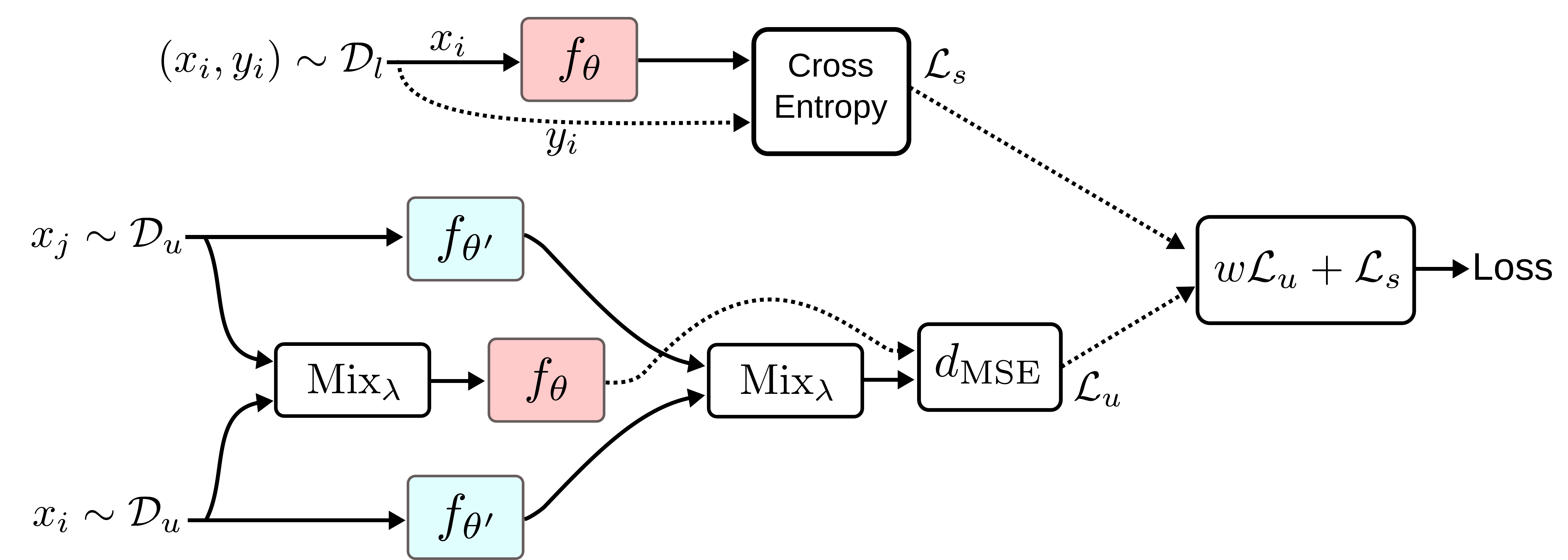}
\caption{\textbf{ICT.} A student model is trained to have consistent predictions at different interpolations of unlabeled data points, where a teacher is used to generate the targets before the Mixup operation.}\label{fig:ict}
\end{figure}

As discussed earlier, the random perturbations are inefficient in high dimensions, given that only a limited subset of the
input perturbations are capable of pushing the decision boundary into low-density regions. VAT and AdD find the adversarial perturbations that will maximize the change in the model's predictions, which
involve multiple forward and backward passes to compute these perturbations. This additional computation can
be restrictive in many cases and makes such methods less appealing.
As an alternative, Verma \etal \cite{verma2019interpolation} propose Interpolation Consistency Training (ICT) as an 
efficient consistency regularization technique for SSL.

Given a MixUp operation \cite{zhang2017mixup}: 
$\operatorname{Mix}_{\lambda}(a, b)=\lambda \cdot a+(1-\lambda) \cdot b$ that outputs an interpolation
between the two inputs with a weight $\lambda \sim \operatorname{Beta}(\alpha, \alpha)$ for $\alpha \in[0, \infty]$.
As shown in \cref{fig:ict}, ICT trains the prediction function $f_{\theta}$ to provide consistent predictions at different interpolations 
of unlabeled data points $x_i$ and $x_j$, where the targets are generated using a teacher model $f_{\theta^{\prime}}$
which is an EMA of $f_{\theta}$:
\begin{equation}
f_{\theta}(\operatorname{Mix}_{\lambda}(x_i, x_j)) \approx
\operatorname{Mix}_{\lambda}(f_{\theta^{\prime}}(x_i), f_{\theta^{\prime}}(x_j))
\end{equation}

The unsupervised objective is to have similar values between the student model's prediction given a mixed input of two unlabeled data points, and the mixed outputs of the teacher model.
\begin{equation}
\mathcal{L}_u = w \frac{1}{|\mathcal{D}_u|} \sum_{x_i, x_j \in \mathcal{D}_u}
 d_{\mathrm{MSE}}(f_{\theta}(\operatorname{Mix}_{\lambda}(x_i, x_j)), 
\operatorname{Mix}_{\lambda}(f_{\theta^{\prime}}(x_i), f_{\theta^{\prime}}(x_j))
\end{equation}

The benefit of ICT compared to random perturbations can be analyzed by considering the mixup operation
as a perturbation applied to a given unlabeled example: $x_i+\delta=\operatorname{Mix}_{\lambda}(x_i, x_j)$,
for a large number of classes and with a similar distribution of examples per class, it is likely that the pair 
of points $(x_i, x_j)$ lie in different clusters and belong to different classes. If one of these two data points
lies in a low-density region, applying an interpolation toward $x_j$ points to a low-density region, which is a good
direction to move the decision boundary toward.

\subsection{Unsupervised Data Augmentation}

Unsupervised Data Augmentation \cite{xie2019unsupervised} uses
advanced data augmentation methods, such as AutoAugment \cite{cubuk2019autoaugment},
RandAugment \cite{cubuk2019randaugment} and Back Translation \cite{edunov2018understanding,sennrich2015improving}, as perturbations 
for consistency training based SSL. Similar to supervised learning, advanced data augmentation methods can also provide extra advantages
over simple augmentations and random noise injection in consistency training, given that; (1) it generates realistic augmented examples, making it safe to encourage the consistency between predictions
on the original and augmented examples, (2) it can generate a diverse set of examples improving the sample efficiency, and (3) it is capable of providing the missing inductive biases for different tasks.

Motivated by these points, Xie \etal \cite{xie2019unsupervised} propose to apply the following
augmentations to generate transformed versions of the unlabeled inputs:
\begin{itemize}
\item \textbf{RandAugment for Image Classification.} Consists of uniformly sampling from the same set of possible
transformations in Python Imaging Library (PIL), without requiring any labeled data to search for a good augmentation strategy.
\item \textbf{Back-translation for Text Classification.} Consists of translating an existing example in language A into another language B, and then translating it back into A to obtain an augmented example.
\end{itemize}

\begin{figure}
\centering
\includegraphics[width = 0.8\textwidth]{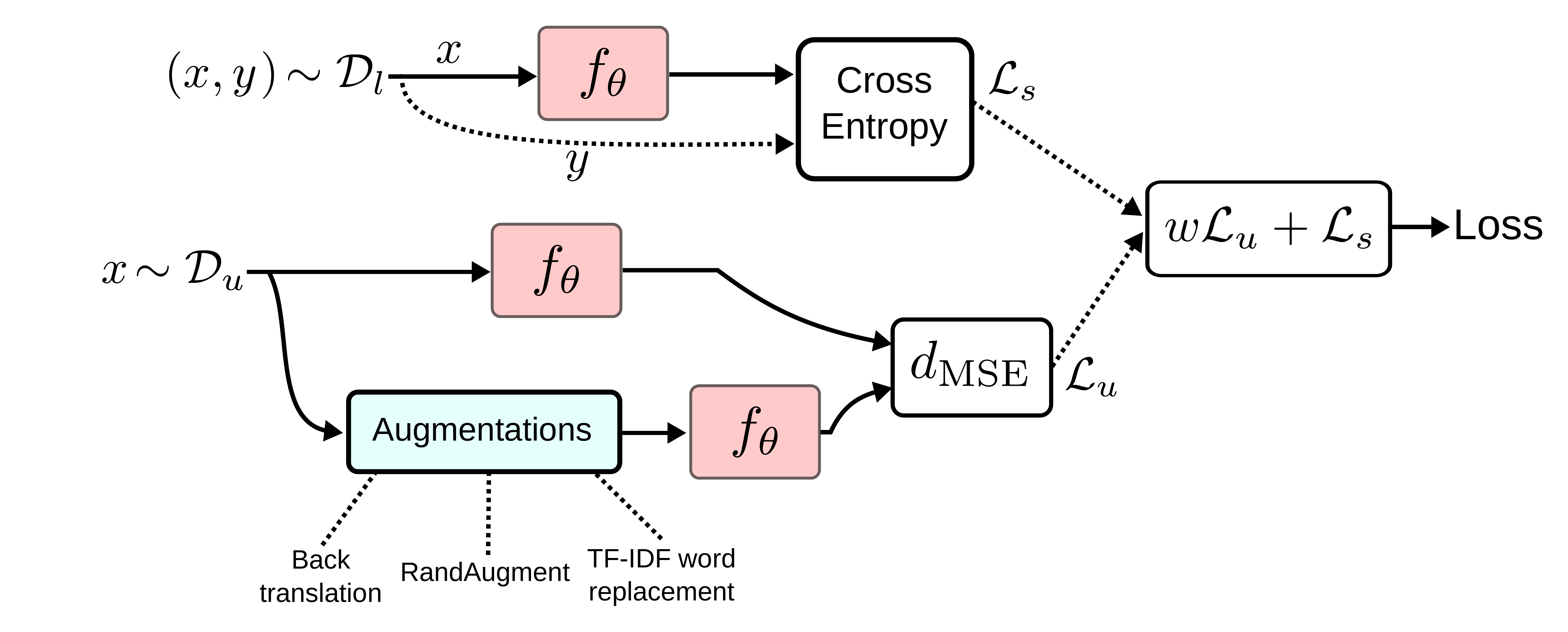}
\caption{\textbf{UDA.} The training procedure consists of computing the supervised loss for the labeled examples and the consistency loss between the two outputs of the augmented and clean input.}
  \label{fig:uda}
\end{figure}

After defining the augmentations to be applied during training, the training procedure (\cref{fig:uda}) is straightforward.
The objective is to have the correct predictions over the labeled set and consistent predictions on the original and augmented examples from the unlabeled set.

%% file: proxylabels.tex
\section{Entropy Minimization}

In the previous section, in a setting where the cluster assumption is maintained, we enforce consistency of predictions
to push the decision boundary into low-density regions to avoid classifying
samples from the same cluster with distinct classes, which is a violation of the cluster assumption.
Another way to enforce this is to encourage the network to make confident (\ie low-entropy) predictions on
unlabeled data regardless of the predicted class, discouraging the decision boundary from passing near data points where it would otherwise be forced to produce low-confidence predictions.
This is done by adding a loss term which minimizes the entropy of the prediction function $f_\theta(x)$.
For a categorical output space with $C$ possible classes, the entropy minimization term \cite{grandvalet2005semi} is:
\begin{equation}
-\sum_{k=1}^{C} f_{\theta}(x)_{k} \log f_{\theta}(x)_{k}
\end{equation}

However, with high capacity models such as neural networks,
the model can quickly overfit to low confident data points by simply outputting large logits, resulting in a model
with very confident predictions \cite{oliver2018realistic}.
On its own, entropy minimization doesn't produce competitive results compared to other SSL methods,
but can produce state-of-the-art results when combined with different approaches.

\section{Proxy-label Methods}

Proxy label methods are the class of SSL algorithms
that produce proxy labels on unlabeled data, using the prediction
function itself or some variant of it without any supervision. These proxy labels
are then used as targets together with the labeled data, providing some additional training information
even if the produced labels are often noisy or weak and do not reflect the ground truth. These methods can be divided mainly
into two groups: self-training, where the model itself produces the proxy labels, and multi-view learning,
where the proxy labels are produced by models trained on different views of the data.

\subsection{Self-training}
In self-training \cite{yarowsky1995unsupervised,scudder1965probability,riloff1996automatically,riloff2003learning},
the small amount of labeled data $\mathcal{D}_l$ is first used to train a prediction function
$f_{\theta}$. The trained model is then used to assign pseudo-labels to the unlabeled data points $x \in \mathcal{D}_u$.
Given an output $f_{\theta}(x)$ for an unlabeled data point $x$ in the form of a probability distribution
over the classes, the pair $(x, \text{argmax}f_{\theta}(x))$ is added to the labeled set if the probability assigned to
its most likely class is higher than a predetermined threshold $\tau$. The process of training the model using
the augmented labeled set, and then set using it to label the remaining of $\mathcal{D}_u$ is repeated until the model is incapable
of producing confident predictions. Other heuristics can be used to decide which proxy labeled examples to retain, such as using the relative confidence
instead of the absolute confidence, where the top $n$ unlabeled samples predicted with the highest
confidence after every epoch are added to the labeled training dataset $\mathcal{D}_l$.
The impact of self-training is similar to that of entropy minimization; in both cases, the network is forced to output more confident predictions.
The main downside of such methods is that the model is unable to correct its own mistakes,
and any biased and wrong classifications can be quickly amplified resulting in confident but erroneous proxy labels on the
unlabeled data points.

Yalnizet \etal \cite{yalniz2019billion} propose to use self-training to improve ResNet-50 \cite{he2016deep} top-1 accuracy and enhance the robustness of the trained model to various perturbations (\eg perturbations used in ImageNet-A, C and P \cite{hendrycks2019benchmarking}). The model is first trained on unlabeled images and their proxy labels, and then fine-tuned on labeled images in the final stage. Instead of using the same model for both proxy labels generation and training, Xie \etal \cite{xie2019self} propose to use the student-teacher setting. In an iterative manner, the teacher model is first trained on the labeled examples and used to generate soft proxy labels on the unlabeled data. The student can then be trained on both the labeled set and the proxy labels while aggressively injecting noise to obtain a more robust model. In the next iteration, the student is considered as a teacher, and a bigger version of EfficientNet \cite{tan2019efficientnet} is used for the student, and the same procedure is repeated up to the largest model.

In addition to image classification, self-training was also successfully applied to a variety of tasks, such as semantic segmentation \cite{babakhin2019semi},
text classification \cite{li2019learning,karamanolakis2019leveraging}, machine translation \cite{sennrich2015improving,he2016dual,cheng2019semi,he2019revisiting} and when learning from noisy data \cite{veit2017learning}.

\paragraph{Pseudo-labeling} \cite{lee2013pseudo,arazo2019pseudo,iscen2019label,shi2018transductive}, similar to self-training, the objective of pseudo-labeling
is to generate proxy labels to enhance the learning process. A first attempt at adapting pseudo-labeling \cite{lee2013pseudo} for deep learning constrained the usage of the proxy labels to a fine-tuning stage after pretraining the network. Shi \etal \cite{shi2018transductive} propose to adapt Transductive SSL \cite{joachims1999transductive,joachims2003transductive,zhang2011fast,wang2016progressive} by treating the labels of unlabeled examples as variables and trying to determine their optimal labels together with the optimal model parameters, by minimizing the proposed loss function through the iterative training process. The generated proxy labels are considered as hard labels for the unlabeled examples, an uncertainty weight is then introduced, with large weights for examples with distant $k$-nearest neighbors in the feature space, in additiont to two loss terms encouraging intra-class compactness and inter-class separation,
 and a consistency term between samples with different perturbations. Iscen \etal \cite{iscen2019label} integrated label-propagation \cite{zhu2002learning,whitney2012bootstrapping,gong2016label}
within pseudo-labeling. The method alternates between training the network on the labeled examples and pseudo-labels and then leveraging the learned representations to build a nearest neighbor graph where label propagation is applied to refine the hard pseudo-labels. They also introduce two uncertainty scores, one for every sample based on the entropy of the output probabilities to overcome the unequal confidence in the predictions, and a per-class scored based class population to deal with class-imbalance. Arazo \etal \cite{arazo2019pseudo} showed that a naive pseudo-labeling overfits to incorrect pseudo-labels due to the so-called confirmation bias, and demonstrate that MixUp \cite{zhang2017mixup} and setting a minimum number of labeled samples per mini-batch are effective regularization techniques for reducing this bias.

\paragraph{Meta Pseudo Labels}
\begin{figure}
\centering
\includegraphics[width = 0.7\textwidth]{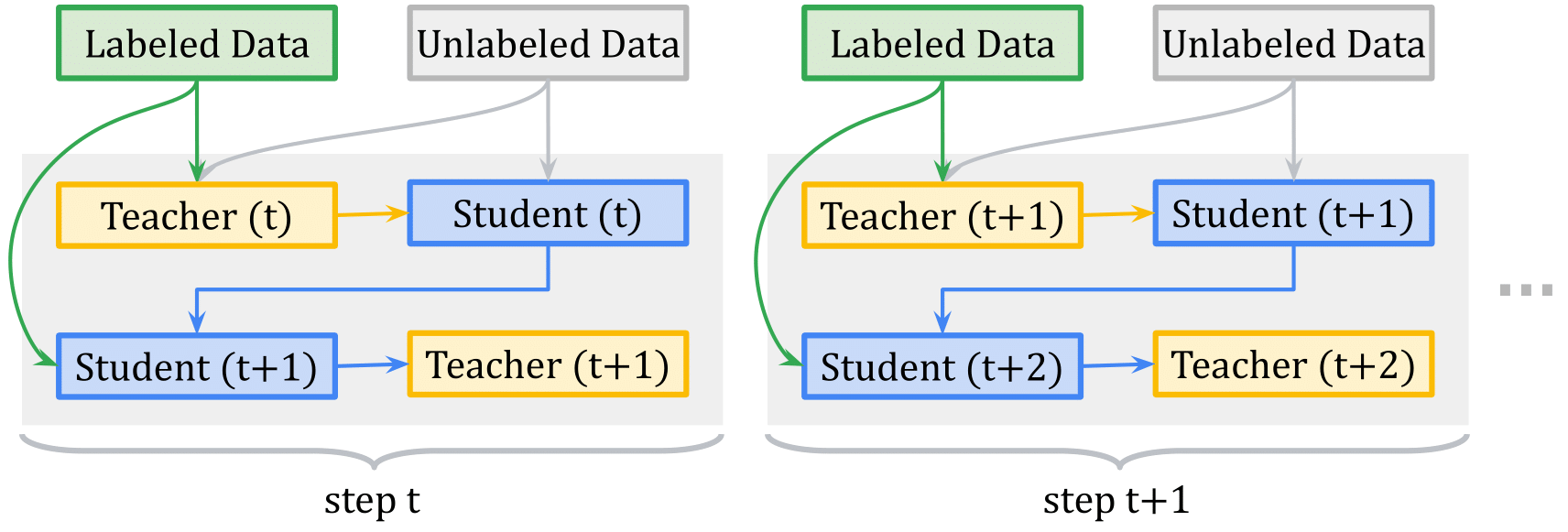}
\caption{The MPL training procedure. At each training iteration, the teacher model is trained along with a student model to set the student's target distributions and adapt to the student's learning state. Image Source: \cite{pham2020meta}.}
\label{fig:mpl}
\end{figure}

Given how important the heuristics used
to select which the proxy labels to add to the training set, where a proper method could lead to a sizable gain.
Pham \etal \cite{pham2020meta} propose to use the student-teacher setting, where the teacher
model is responsible for producing the proxy labels based on an efficient meta-learning algorithm called Meta Pseudo Labels (MPL),
which encourages the teacher to adjust the target distributions 
of training examples in a manner that improves the learning of the student model. The teacher is updated by policy gradients computed 
by evaluating the student model on a held-out validation set.

A given training step of MPL consists of two phases (\cref{fig:mpl}):
\begin{itemize}
\item \textbf{Phase 1:} The Student learns from the Teacher. In this phase, given a single input example $x \in \mathcal{D}_l$, the teacher $f_{\theta^{\prime}}$ produces a target class-distribution to train the student $f_{\theta}$, where
the pair $(x, f_{\theta^{\prime}}(x))$ is shown to the student to update its parameters by back-propagating from the cross-entropy loss.
\item \textbf{Phase 2:} The Teacher learns from the Student's Validation Loss. After the student updates its parameters in first step,
its new parameter $\theta(t+1)$ are evaluated on an example $(x_{val},y_{val})$ from the held-out validation dataset
using the cross-entropy loss. Since the validation loss
depends on $\theta^{\prime}$ via the first step, this validation cross-entropy loss is also a function of the teacher's weights $\theta^{\prime}$.
This dependency allows us to compute the gradients of the validation loss with respect to the teacher's weights, and then update $\theta^{\prime}$
to minimize the validation loss using policy gradients.
\end{itemize}

While the student's performance allows the teacher to adjust and adapt to the student's learning state, this signal alone
is not sufficient to train the teacher since when the teacher has observed enough evidence to produce meaningful target
distributions to teach the student, the student might have already entered a bad region of parameters. To overcome this, the teacher is 
also trained using the pair of labeled data points from the held-out validation set.

\subsection{Multi-view training}

Multi-view training (MVL) \cite{kumar2011co,zhao2017multi}
utilizes multi-view data that are very common in real-world applications, where 
different views can be collected by different measuring methods (\eg color information and texture information for images)
or by creating limited views of the original data. In such a setting, MVL aims to learn a distinct prediction function $f_{\theta_i}$
to model a given view $v_{i}(x)$ of a data point $x$, and jointly optimize all the functions to improve the generalization performance.
Ideally, the possible views complement each other so that the produced models can collaborate in improving each other's performance.

\subsubsection{Co-training}

Co-training \cite{blum1998combining} requires that each data point $x$ can be represented using
two conditionally independent views $v_1(x)$ and $v_2(x)$, and that each view is sufficient to train a good model.
After training two prediction functions $f_{\theta_1}$ and $f_{\theta_2}$ on a specific view on the labeled set $\mathcal{D}_l$.
We start the proxy labeling procedure. At each iteration, an unlabeled data point is added to the training
set of the model $f_{\theta_i}$ if the other model $f_{\theta_j}$ outputs a confident prediction with a probability higher
than a threshold $\tau$. This way, one of the models provides newly labeled examples where the other model is uncertain.
Co-training has been combined with deep learning for some applications, such as object recognition \cite{cheng2016semi} by utilizing RGB-D data, with RGB and depth
as the two views used to train the two models, or for combining multi-modal data \cite{ardehaly2017co} (\ie image and text) by training each model on a given
modality and use it to provide pseudo-labels for other models.
However, in many cases, the data have only one view rather than two, in this instance, different learning algorithms or different parameter
configurations to learn two different classifiers can be employed.
The two views $v_1(x)$ and $v_2(x)$ can also be generated by injecting noise or by applying different augmentations,
for example, Qiao \etal \cite{qiao2018deep} used adversarial perturbations to produce new views for deep co-training
for image classification, where the models are encouraged to have the same predictions on $\mathcal{D}_l$ but make different 
errors when they are exposed to adversarial attacks.

\paragraph{Democratic Co-training} \cite{zhou2004democratic}. An extension of Co-training, consists of replacing the different views of the input data with a number of models with different architectures
and learning algorithms, which are first trained on the labeled examples. The trained models are then used to label a given example $x$ if a majority of models confidently agree on its label.

\subsubsection{Tri-Training}
Tri-training \cite{zhou2005tri} tries to overcome the lack of data with multiple views and reduce the
bias of the predictions on unlabeled data produced with self-training
by utilizing the agreement of three independently trained models instead of a single model.
First, the labeled data $\mathcal{D}_l$ is used to train three prediction functions:
$f_{\theta_1}$, $f_{\theta_2}$ and $f_{\theta_3}$.
An unlabeled data point $x \in \mathcal{D}_u$ is then added to the supervised training set of the function $f_{\theta_i}$ if the other two models
agree on its predicted label. The training stops if no data points are being added to any of the models' training sets.
Tri-training requires neither the existence of multiple views nor unique learning algorithms, making it more generally applicable.
Using Tri-training with neural networks can be very expensive, requiring predictions for each one of the three models on all the unlabeled data. Ruder \etal \cite{ruder2018strong} propose to sample a limited number of unlabeled data points at each training epoch,
the candidate pool size is increased as the training progresses and the models become more accurate.

\paragraph{Multi-task tri-training} \cite{ruder2018strong}
can also be used to reduce the time and sample complexity, where all three models
share the same feature-extractor with model-specific classification layers. This way, the models are trained jointly
with an additional orthogonality constraint on two of the three classification layers to be added to loss term, to avoid
learning similar models and falling back to the standard case of self-training. Tri-Net \cite{dong2018tri} also falls in this category, with a shared module for joint learning and three output modules for tri-training, in addition to utilizing output smearing \cite{breiman2000randomizing} to initialize these modules. After the proxy labeling iteration, a fine-tuning stage is conducted on the labeled data to augment diversity and eliminate unstable and suspicious pseudo-labeled data.

\paragraph{Cross-View Training} 

\begin{figure}
\centering
\includegraphics[width = 0.6\textwidth]{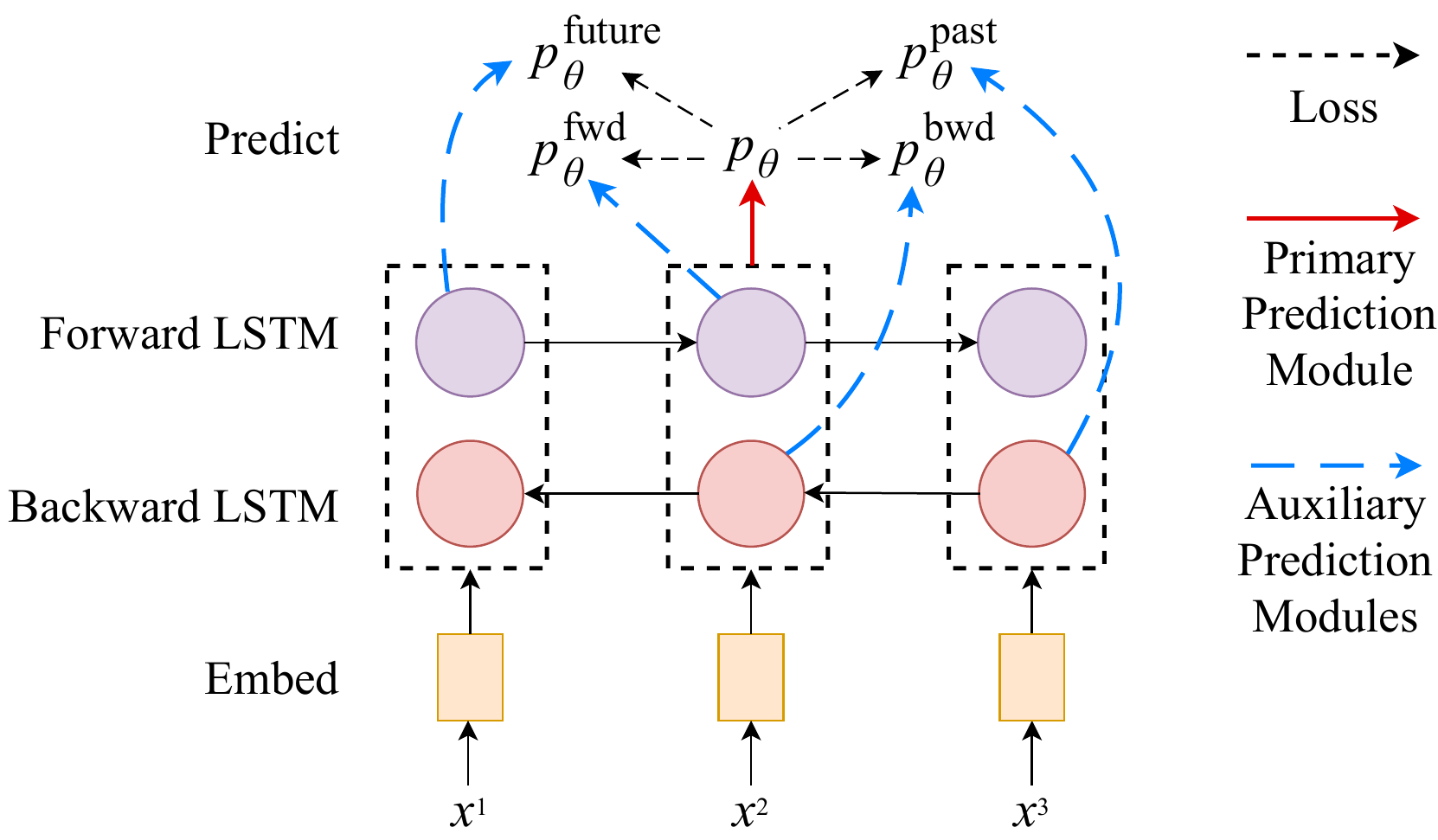}
\caption{\textbf{Cross-view Training.} An example of auxiliary student prediction modules. Each student sees a restricted view of the input.  For instance, the \textit{forward} prediction module does not see any context to the right of the current token when predicting that tokens label. Image Source: \cite{clark2018semi}}\label{fig:cvt}
\end{figure}

In self-training, the model plays a dual role of a teacher and a student, producing the predictions it is being trained on, resulting in very moderate performance gains. As a solution, and taking inspiration from multi-view learning and consistency training, Clark \etal \cite{clark2018semi} propose Cross-View Training, where the model is trained to produce consistent predictions across different views of the inputs. Instead of using a single model as a teacher and a student,
they propose to use a shared encoder, and then add auxiliary prediction modules that transform the encoder representations into predictions, these modules are then divided into auxiliary student modules and a primary teacher module. The input to each student prediction module is a subset of the model's intermediate representations corresponding to a restricted view of the input, such as feeding one of the student only the forward LSTM from a given Bi-LSTM layer, so it makes predictions without seeing any tokens to the right of the current one (\cref{fig:cvt}). The primary teacher module in trained only on labeled examples, and is responsible of generating the pseudo-labels taking as input the full view of the unlabeled inputs, the students are trained to have consistent predictions with the teacher module. Given an encoder $e$, a teacher module $t$ and $K$ student modules $s_i$ with $i \in [0, K]$, where each student receives a limited view of the input, the training objective is written as follows:
\begin{equation}
\mathcal{L} = \mathcal{L}_u  + \mathcal{L}_s = 
\frac{1}{|\mathcal{D}_u|} \sum_{x\in \mathcal{D}_u} \sum_{i = 1}^K d_{\mathrm{MSE}}(t(e(x)), s_i(e(x))) +
\frac{1}{|\mathcal{D}_l|} \sum_{x, y \in \mathcal{D}_l} \mathrm{H}(t(e(x)), y)
\end{equation}

Cross-view training takes advantage of unlabeled data by improving the encoder's representation learning.  The student prediction modules can learn from the teacher module predictions because this primary module has a better, unrestricted view of the inputs. As the student modules learn to make accurate predictions despite their restricted views of the input, they improve the quality of the representations produced by the encoder. Which, in turn, improves the full model, which uses the same shared representations.

%% file: holostic.tex
\section{Holistic Methods}

An emerging line of work in SSL is a set of holistic approaches that try to unify the current dominant methods in SSL
in a single framework, achieving better performances.

\subsection{MixMatch}
\begin{figure}
\centering
\includegraphics[width = 0.8\textwidth]{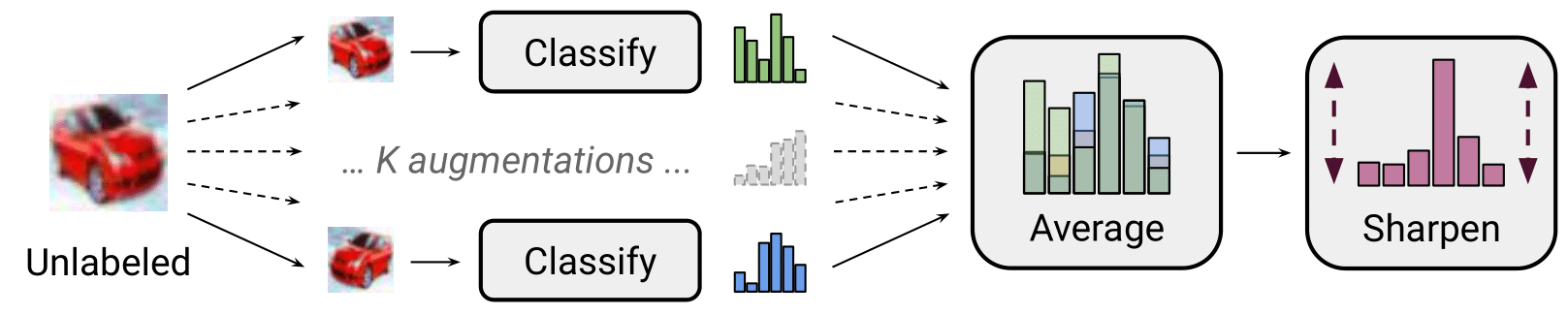}
\caption{\textbf{MixMatch.} The procedure of label guessing process used in MixMatch, taking as input a batch of unlabeled examples, and 
outputting a batch of $K$ augmented version of each input, with a corresponding sharpened proxy labels. Image Source: \cite{pham2020meta}.}
  \label{fig:mixmatch}
\end{figure}

Berthelot \etal \cite{berthelot2019mixmatch} propose a \textit{holistic} approach which gracefully unifies
ideas and components from the dominant paradigms for SSL, resulting in an algorithm that is greater than the sum of its parts
and surpasses the performance of the traditional approaches.

MixMatch takes as input a batch from the labeled set $\mathcal{D}_l$ containing pairs of inputs and their corresponding one-hot targets,
a batch from the unlabeled set $\mathcal{D}_u$ containing only unlabeled data, and a set of hyperparameters: the sharpening softmax temperature $T$,
the number of augmentations $K$, and the Beta distribution parameter $\alpha$ for MixUp. Producing a batch of augmented labeled examples
and a batch of augmented unlabeled examples with their proxy labels. These augmented examples can then be used to
compute the losses and train the model. Precisely, MixMatch consists of the following steps:
\begin{itemize}
\item \textbf{Step 1: Data Augmentation.} Using a given transformation, a labeled example $x \in \mathcal{D}_l$ from the labeled batch is transformed, producing its augmented versions $\tilde{x}$. For an unlabeled example $x \in \mathcal{D}_u$,
the augmentation function is applied $K$ times, resulting in $K$ augmented
versions of the unlabeled examples {$\tilde{x}_1$, ..., $\tilde{x}_K$}.
\item \textbf{Step 2: Label Guessing.} The second step consists of producing proxy labels for the unlabeled examples.
First, we generate the predictions for the $K$ augmented versions of 
each unlabeled example using the predictions function $f_\theta$. The $K$ predictions are then averaged together, obtaining 
a proxy or a pseudo label $\hat{y} = 1/K \sum_{k=1}^{K}(\hat{y}_k)$ for each one of the augmentations of the unlabeled example $x$:
{($\tilde{x}_1, \hat{y}$), ..., ($\tilde{x}_K, \hat{y}$)}.
\item \textbf{Step 3: Sharpening.} To push the model to produce confident predictions and minimize the entropy of the output distribution, the generated
proxy labels $\hat{y}$ in step 2 in the form of a probability distribution over $C$ classes are sharpened by adjusting the temperature
of the categorical distribution, computed as follows where $(\hat{y}^u)_k$ refers to the probability of class $k$ out of $C$ classes:
\begin{equation}
(\hat{y})_k = (\hat{y})_k^{\frac{1}{T}} / \sum_{k=1}^{C} (\hat{y})_k^{\frac{1}{T}}
\end{equation}
\item \textbf{Step 4 MixUp.} The previous steps resulted in two new augmented batches, a batch $\mathcal{L}$ of augmented labeled
examples and their target, and a batch $\mathcal{U}$ of augmented unlabeled examples and their sharpened proxy labels. Note that the size
of $\mathcal{U}$ is $K$ times larger than the original batch given that each example $x \in \mathcal{D}_u$ is replaced by its $K$
augmented versions. In the last step, we mix these two batches. First, a new batch merging both batches is created
$\mathcal{W}=\text{Shuffle}(\text{Concat}(\mathcal{L}, \mathcal{U}))$. $\mathcal{W}$ is then
divided into two batches: $\mathcal{W}_1$ of the same size as $\mathcal{L}$ and $\mathcal{W}_2$ of the same
size as $\mathcal{L}$. Using the Mixup operation that is slightly adjusted so that the mixed
example is closer the labeled examples, the final step is to create new labeled and unlabeled batches by mixing the produced batches together
using Mixup as follows:
\end{itemize}
\begin{equation}
\mathcal{L}^{\prime}=\operatorname{MixUp}(\mathcal{L}, \mathcal{W}_1)
\end{equation}
\begin{equation}
\mathcal{U}^{\prime}=\operatorname{MixUp}(\mathcal{U}, \mathcal{W}_2)
\end{equation}

After creating two augmented batches $\mathcal{L}^{\prime}$ and $\mathcal{U}^{\prime}$ using MixMatch,
we can then train the model using the standard SSL losses by computing the CE loss for the supervised loss, and
the consistency loss for the unsupervised loss using the augmented batches as follows:
\begin{equation}
\mathcal{L} = \mathcal{L}_s + w \mathcal{L}_u =\frac{1}{|\mathcal{L}^{\prime}|} \sum_{x, y \in \mathcal{L}^{\prime}} \mathrm{H}(y, f_\theta(x))) +
w \frac{1}{|\mathcal{U}^{\prime}|} \sum_{x, \hat{y} \in \mathcal{U}^{\prime}} d_{\mathrm{MSE}}(\hat{y}, f_{\theta}(x))
\end{equation}

\subsection{ReMixMatch}

Berthelot \etal \cite{berthelot2019remixmatch} propose to
improve MixMatch by introducing two new techniques: \textbf{distribution alignment} and \textbf{augmentation anchoring}.
Distribution alignment encourages the marginal distribution of predictions on unlabeled data
to be close to the marginal distribution of ground-truth labels. Augmentation anchoring feeds multiple strongly
augmented versions of the input into the model, encouraging each output to be close to the prediction
for a weakly-augmented version of the same input.

\begin{figure}
\centering
\includegraphics[width = 0.9\textwidth]{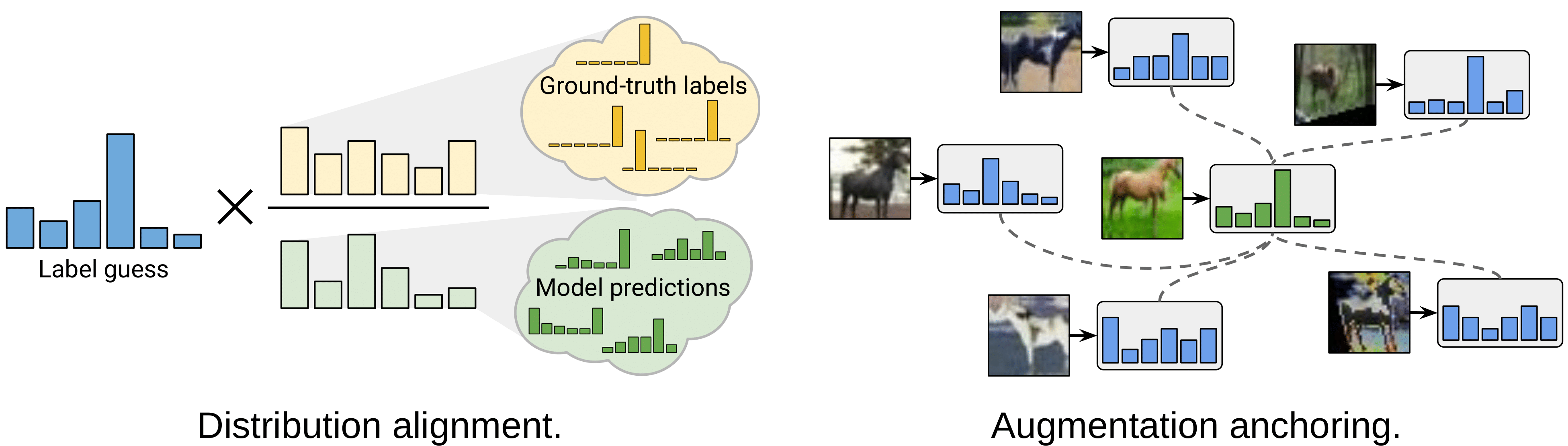}
\caption{\textbf{ReMixMatch.} \textit{Left.} Distribution alignment adjusts the guessed labels distributions to match the ground-truth class distribution divided by the average model predictions on $\mathcal{D}_u$. \textit{Right.} Augmentation anchoring uses the prediction obtained using a weakly augmented image as targets for a strongly augmented version of the same image. Image Source: \cite{berthelot2019remixmatch}.}
  \label{fig:remixmatch}
\end{figure}

\noindent \textbf{Distribution alignment.} In order to force that the aggregate of predictions on unlabeled data matches
the distribution of the provided labeled data. Over the course of training, a running average $\tilde{y}$ of the model's predictions
on unlabeled data is maintained over the last 128 batches. For the marginal class distribution $p(y)$, it is estimated based on the labeled
examples seen during training. Given a prediction $f_{\theta}(x)$ on the unlabeled example $x$, the output probability distribution
is aligned as follows: $f_{\theta}(x) = \text { Normalize }(f_{\theta}(x) \times p(y) / \tilde{y})$.

\noindent \textbf{Augmentation Anchoring.} MixMatch uses a simple flip-and-crop augmentation strategy, ReMixMatch
replaces the weak augmentations with strong augmentations learned using a control theory based augmentation strategy following AutoAugment. With such augmentations, the model's prediction for a weakly augmented unlabeled image is used as a proxy label for many strongly augmented versions of the same image in a standard cross-entropy loss.

For training, MixMatch is applied to the unlabeled and labeled batches, with the application of distribution alignment and replacing the $K$ weakly
augmented example with a strongly augmented example, in addition to using the weakly augmented examples for predicting proxy labels
for the unlabeled strongly augmented examples. With two augmented batches $\mathcal{L}^{\prime}$ and $\mathcal{U}^{\prime}$, the
supervised and unsupervised losses are computed both using the cross-entropy loss as follows:
\begin{equation}
\mathcal{L} = \mathcal{L}_s + w \mathcal{L}_u =
\frac{1}{|\mathcal{L}^{\prime}|} \sum_{x, y \in \mathcal{L}^{\prime}} \mathrm{H}(y, f_\theta(x))) +
w \frac{1}{|\mathcal{U}^{\prime}|} \sum_{x, \hat{y} \in \mathcal{U}^{\prime}} \mathrm{H}(\hat{y}, f_\theta(x)))
\end{equation}

In addition to these losses, the authors add a self-supervised loss. First, a new unlabeled batch
$\hat{\mathcal{U}}^{\prime}$ of examples is created by rotating all of the examples with an angle $r \sim\{0,90,180,270\}$. The rotated
examples are then used to compute a self-supervised loss, where the classification layer on top of the model predicts the correct applied
rotation, in addition to the cross-entropy loss over the rotated examples:
\begin{equation}
\mathcal{L}_{SL} = w^{\prime}
\frac{1}{|\hat{\mathcal{U}}^{\prime}|} \sum_{x, \hat{y} \in \hat{\mathcal{U}}^{\prime}}
\mathrm{H}(\hat{y}, f_\theta(x))) + \lambda
\frac{1}{|\hat{\mathcal{U}}^{\prime}|} \sum_{x \in \hat{\mathcal{U}}^{\prime}}
\mathrm{H}(r, f_\theta(x)))
\end{equation}

\subsection{FixMatch}
\begin{figure}
\centering
\includegraphics[width = 0.6\textwidth]{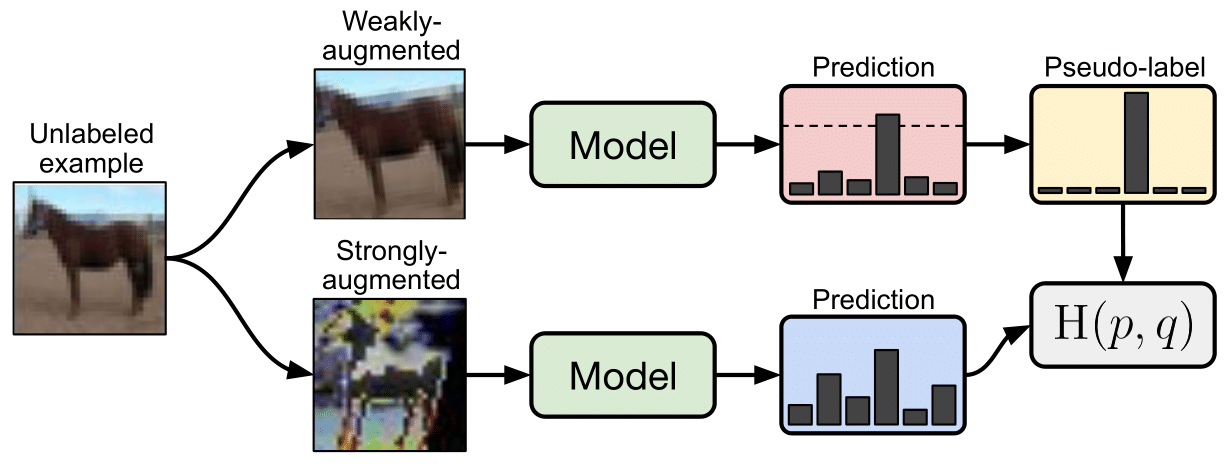}
\caption{\textbf{FixMatch.} The model prediction on a weakly augmented input is considered as target if the maximum output class probability is above threshold, this target can then be used to train the model on a strongly augmented version of the same input using standard cross-entropy loss. Image Source: \cite{berthelot2019remixmatch}.}
\label{fig:fixmatch}
\end{figure}

FixMatch \cite{sohn2020fixmatch} proposes a simple SSL algorithm that combines consistency regularization and pseudo-labeling.
In FixMatch (\cref{fig:fixmatch}), both the supervised and unsupervised losses are computed using a cross-entropy loss.
For labeled examples, the provided targets are used. For unlabeled examples $x \in \mathcal{D}_u$, a weakly augmented version is first computed using weak augmentation function $A_w$. As in self-training, the predicted label is then considered as a proxy label if the highest class probability is greater than a threshold $\tau$. With a proxy label, $K$ strongly augmented examples are generated
using a strong augmentation function $A_s$. We then assign to these augmented versions the proxy label obtained with the weakly
labeled version. The unsupervised loss can be written as follows:
\begin{equation}
\mathcal{L}_u = w \frac{1}{K |\mathcal{D}_u|} \sum_{x \in \mathcal{D}_u} \sum_{i=1}^{K}
\mathrm{1}(\max (f_\theta(A_w(x))) \geq \tau) \mathrm{H} (f_\theta(A_w(x)), f_\theta(A_s(x)))
\end{equation}

\noindent \textbf{Augmentations.} Weak augmentations consist of a standard flip-and-shift augmentation strategy.
Specifically, the images are flipped horizontally with a probability of 50\% on all datasets except SVHN, in addition to randomly translating
images by up to 12.5\% vertically and horizontally. For the strong augmentations, RandAugment and CTAugment \cite{berthelot2019remixmatch} are used
where a given transformation (\eg color inversion, translation, contrast adjustment, etc.) is randomly selected for each sample in a batch
of training examples, and the amplitude of the transformation is a hyperparameter that is optimized during training.

Other important factors in the FixMatch are the usage of Adam optimizer \cite{kingma2014adam},
weight decay regularization and the learning rate schedule, where the authors propose to use a cosine learning rate decay with a decay of 
$\eta \cos (\frac{7 \pi t}{16 T})$, where $\eta$ is the initial learning rate, $t$ is the current training step, and $T$ is the total number of training iterations.

%% file: generativemodels.tex
\section{Generative Models}

In unsupervised learning, we are provided with samples $x$ drawn i.i.d. from an unknown data distribution with
density $p(x)$, and the objective is to estimate this density. Supervised learning, on the other hand, consists of estimating a functional relationship between the inputs $x$ and the labels $y$ with the goal of minimizing
the functional of the joint distribution $p(x, y)$ \cite{chapelle2009semi}. Classification can be treated as a special case of estimating $p(x, y)$,
where we are only interested in the conditional distributions $p(y|x)$, without the need to estimate the input distribution $p(x)$ since $x$ will always be given at prediction time. Semi-supervised learning with generative models can be viewed as either an extension of supervised learning,
classification in addition to information about $p(x)$ provided by $\mathcal{D}_u$, or as an extension of unsupervised learning,
clustering in addition to the provided labels from $\mathcal{D}_l$. In this section, we explore some generative approaches for deep SSL.

\subsection{Variational Autoencoders for SSL}
Variational Autoencoders (VAEs) \cite{kingma2013auto,doersch2016tutorial} have emerged as one of the most popular approaches
to unsupervised learning of complicated distributions. A standard VAE is an autoencoder trained with a reconstruction objective between the
inputs and their reconstructed versions, in addition to a variational objective term that attempts to learn a latent space that roughly follows
a unit Gaussian distribution, this objective is implemented as the KL-divergence between the latent space and the standard Gaussian.
With an input $x$, the conditional distribution $q_\phi(z|x)$ modeled by an encoder, the standard Gaussian distribution
$p(z)$ and the reconstructed input $\hat{x}$ generated using a decoder $p_\theta(x|z)$. The parameters $\phi$ and $\theta$ are
trained to minimize the following objective:
\begin{equation}
\mathcal{L}=d_{\mathrm{MSE}}(x, \hat{x})+ d_{\mathrm{KL}}(q_\phi(z|x), p(z))
\end{equation}

\subsubsection{Variational Autoencoder}
Kingma \etal \cite{kingma2014semi} expanded the work on variational generative techniques \cite{kingma2013auto,rezende2014stochastic} for SSL, that exploit generative descriptions of the data to improve upon the classification performance that would be obtained using the labeled data alone.

\paragraph{Standard VAEs for SSL (M1 Model)}
The first model consists of an unsupervised pretraining stage, in which the VAE is trained using the labeled and unlabeled examples. Using a fully trained VAE, the observed labeled
data $x\in \mathcal{D}_l$ are transformed into the latent space defined by $z$, the standard supervised task can then be solved using $(z, y)$ where $y$ are the labels of $x$. With this approach, the classification can be performed in a lower dimensional space since the dimensionality of the latent variables $z$ is much less than that of the observations. These low dimensional embeddings are more easily separable since the latent space is formed by independent Gaussian posteriors parameterized by an encoder, built by a sequence of non-linear transformations of the inputs.

\paragraph{Extending VAEs for SSL (M2 Model)}
In the M1 model, the labels of $\mathcal{D}_l$ were ignored when training the VAE. With the second model, the labels are also used during training. If the class labels are not available, $y$ is treated as a latent variable in addition to the latent variable $z$. The network in this case contains three components, $q_\phi(y|x)$ modeled by a classification network, $q_\phi(z|y,x)$ modeled by an encoder, and $p_\theta(x|y,z)$ modeled by a decoder, with parameters $\phi$ and $\theta$. The training is similar to a standard VAE with the addition of the posterior on $y$ and loss terms to train $q_\phi(y|x)$ if the labels are available. The distribution $q_\phi(y|x)$ can then be used at test time to get the predictions on unseen data.

\paragraph{Stacked VAEs (M1+M2 Model)}
The two previous models can be concatenated to form a joint model. In this case, the model M1 is first trained to obtain the latent variables $z_1$, the model M2 then uses the latent variables $z_1$ from model M1 as new representations of the data as opposed to raw values $x$. The final model can be described as follows:
\begin{equation}
p_{\theta}(x, y, z_1, z_2)= p(y)p(z_2)p_{\theta}(z_1|y, z_2) p_{\theta}(x|z_1)
\end{equation}

\subsubsection{Variational Auxiliary Autoencoder}
Variational Auxiliary Autoencoder \cite{maaloe2016auxiliary,ranganath2016hierarchical} extends the variational distribution with auxiliary variables $a$: $q(a,z|x)=q(z|a,x) q(a|x)$, such that the marginal distribution $q(z|x)$ can fit more complicated posteriors $p(z|x)$ while improving the flexibility of inference. In order to have an unchanged generative model $p(x|z)$, it is required that the joint mode $p(x,z,a)$ gives back the original $p(x,z)$ under marginalization over $a$, thus $p(x,z,a) =p(a|x,z)p(x,z)$, with $p(a|x,z) \neq p(a)$ to avoid falling back to the original VAE model.

In SSL, to incorporate the class information, an additional latent variable $y$ is introduced, the generative model become $p(y) p(z) p(a|z,y,x) p(x|y,z)$, with $a$, $y$, $z$ as the auxiliary variable, class label, and latent features respectively. In this case, the auxiliary unit $a$ introduces a latent feature extractor to the inference model giving a richer mapping between $x$ and $y$. The resulting model is parametrized by 5 neural networks: 1) an auxiliary inference model $q(a|x)$, 2) a latent inference model $q(z|a,y,x)$, 3) a classification model $q(y|a,x)$, 4) a generative model $p(a|.)$, and 5) a generative model $p(x|.)$, which are trained on both a generative and discriminative tasks simultaneously.

\subsubsection{Infinite Variational Autoencoder}

Another variation of VAEs for SSL is Infinite Variational Autoencoder \cite{ehsan2017infinite}, to overcome the limitation of VAEs of having a fixed dimension of the latent space and a fixed number of parameters in the generative model in advance, in which the capacity of the model must be chosen a priori with some foreknowledge of the training data characteristics. Infinite VAE solves this by producing an infinite mixture of autoencoders capable of growing with the complexity of the data to best capture its intrinsic structure. After training the generative model using unlabeled data, this model can then be combined with the available labeled data to train a discriminative model, which is also a mixture of experts, for classification. For a given test example $x$, each discriminative expert produces a tentative output that is then weighted by the generative model. As such, each discriminative expert learns to perform better with instances that are more structurally similar from the generative model's perspective. With a higher modeling capability, the infinite VAE is able to capture the distribution of the unlabeled data more accurately. Therefore, it provides a generative model that allows the discriminative model, which is trained based on its output, to be more effectively learned using a small number of samples.

\subsection{Generative Adversarial Networks for SSL}
\begin{figure}
\centering
\includegraphics[width = 0.7\textwidth]{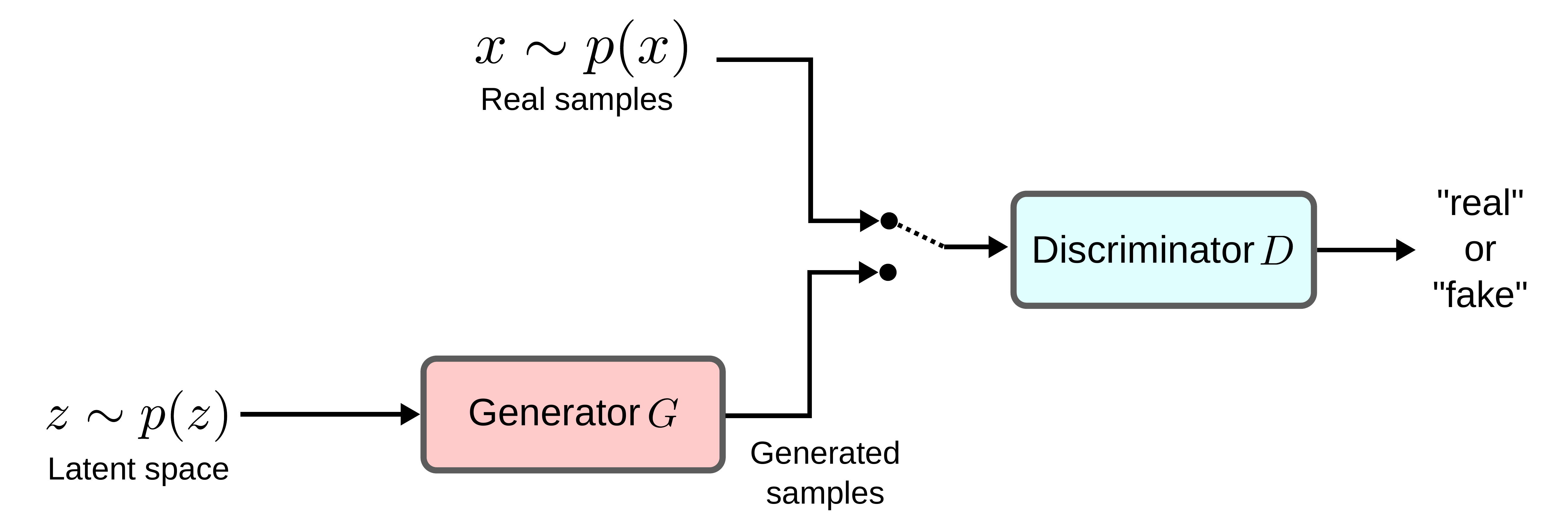}
\caption{\textbf{GAN framework.} During training, the discriminator $D$ alternates between receiving real samples from the data distribution $p(x)$, with
the goal of correctly classifying them as real, \ie $D(x)=1$, and generated samples $G(z)$ with the aim of correctly classifying them as fake, \ie $D(G(z))=0$, while competing with the generator, trying to generate real-looking samples to fool the discriminator, \ie $D(G(z))=1$.}
  \label{fig:gan}
\end{figure}

A Generative Adversarial Network (GAN) \cite{goodfellow2014generative} (\cref{fig:gan}) consists of a generator network $G$ and a discriminator network $D$. The generator receives a latent variable 
$z \sim p(z)$ sampled from the prior distribution $p(z)$ and maps to the input space. The discriminator takes an input, either coming from the real data $p(x)$ or
generated by $G$ and outputs the probability of the input being from either $G$ or the real data distribution $p(x)$, represented with an empirical distribution $\mathcal{D}$. The standard training procedure of GANs minimizes two objectives by alternating between training the discriminator $D$ and the generator $G$:
\begin{equation}
\begin{split}
\mathcal{L}_{D}&=\max_{D}\mathbb{E}_{x \sim p(x)}[\log D(x)]+\mathbb{E}_{z \sim p(z)}[1-\log D(G(z))] \\
\mathcal{L}_{G}&=\min_{G}-\mathbb{E}_{z \sim p(z)}[\log D(G(z))]
\end{split}
\end{equation}
where $p(z)$ is usually chosen as a standard normal distribution. Other formulations have been proposed to improve and stabilize the training procedure, such as the hinge-loss version of the adversarial loss \cite{lim2017geometric,tran2017deep} and Wassertein GAN (WGAN) \cite{arjovsky2017wasserstein}. Which are subsequently
improved in several ways \cite{zhang2018self,miyato2018spectral,zhang2019consistency}, such as using spectral normalization \cite{miyato2018spectral} on both the generator and the discriminator, or consistency regularization on the discriminator \cite{zhang2019consistency}.

\subsubsection{CatGAN}

Categorical generative adversarial networks (CatGAN) \cite{springenberg2015unsupervised} consist of combining both the generative and the discriminative perspectives within the training procedure. The discriminator $D$ in this case plays the role of $C$ classifiers and is trained to maximize the mutual information between the inputs $x$ and the predicted labels for a number of $C$ unknown classes. To aid these classifiers in their task of discovering categories that generalize well to unseen data, and avoid overfitting to spurious correlations in the data, the adversarial generative network comes into play and provides the examples the discriminator must become robust to.

The traditional two-player game in the GAN framework can be extended to CatGAN by having a discriminator that assign all examples to one the $C$ classes instead of a probability of $x$ belonging to $p(x)$, while staying uncertain of the class assignments for the generated samples by $G$. After training such a classifier-generator pair where the discovered $C$ classes coincide with the classification problem we are interested in, the classifier can then be used during inference being trained only on unlabeled data.

CatGAN objective dictates three requirements for the discriminator and two requirements that the generator should fulfilled:
\begin{itemize}
\vspace{-0.8em}
\item Discriminator requirements: should (1) be certain of class assignment for samples from $p(x)$, (2) be uncertain of assignment for generated samples, and (3) by assuming a uniform prior $p(y)$ over classes, all classes must be distributed equally.
\item Generator requirements: should (1) generate samples with highly certain class assignments, and (2) similar to the discriminator, equally distribute samples across all classes.
\end{itemize}

In order to have the output class distribution $D(x) = p(y|x, D)$ to be highly peaked where $D$ is certain about the class assignment, the entropy $\mathrm{H}(D(x))$ of the class distribution must be low. For the generated samples $D(G(z)) = p(y|G(z), D)$, the predictions should be highly uncertain with a uniform class distribution, in this case, the entropy $\mathrm{H}(D(G(z)))$ must be high. The first two requirements can then be enforced by simply minimizing $\mathrm{H}(D(x))$  and maximizing the $\mathrm{H}(D(G(z)))$. To meet the third requirement that all classes should be used equally, the entropy of the marginal class distribution as measured empirically for both $D$ and $G$ needs to be maximized:
\begin{equation}
\begin{aligned}
&\mathrm{H}_{\mathcal{D}}= \mathrm{H}\left(\frac{1}{N} \sum_{i=1}^{N} D(x_i)\right) \\
&\mathrm{H}_{G} \approx \mathrm{H}\left(\frac{1}{M} \sum_{i=1}^{M} D(G(z_i))\right)
\end{aligned}
\end{equation}

Combining these requirements, CatGAN objective for the discriminator and the generator is:
\begin{equation}
\begin{split}
\mathcal{L}_{D}&=\max_{D}-\mathbb{E}_{x \sim p(x)}[\mathrm{H}(D(x))]+\mathbb{E}_{z \sim p(z)}[\mathrm{H}(D(G(z)))]+\mathrm{H}_{\mathcal{D}} \\
\mathcal{L}_{G}&=\min_{G}\mathbb{E}_{z \sim p(z)}[\mathrm{H}(D(G(z)))]-\mathrm{H}_{G}
\end{split}
\end{equation}

In SSL, if the input $x$ comes from the labeled set $\mathcal{D}_l$ with a label $y$ in the form of a one-hot vector, the discriminator $D$ is trained with a cross-entropy loss in addition to $\mathcal{L}_{D}$:
\begin{equation}
\mathcal{L}_{D} + \lambda \mathbb{E}_{(x, y) \sim p(x)_l}[-y \log G(x)]
\end{equation}
where $\lambda$ is a cost weighting term.

\subsubsection{DCGAN}
Another way of using GANs for SSL is to leverage the unlabeled examples to learn good and transferable intermediate representations, which can then be used on a variety of supervised learning tasks such as image classification based on a small labeled set $\mathcal{D}_l$. 
Radford \etal \cite{radford2015unsupervised} propose to build good image representations by training GANs, and later reusing parts of the generator and discriminator networks as feature extractors for supervised tasks. The authors propose Deep Convolutional GANs (DCGAN), a class of architectures with a set of constraints on the architectural topology of convolutional GANs to be able to scale them while maintaining a stable training in most settings, such as replacing polling layers with strided convolutions for the discriminator, fractional-strided convolutions for generator, using batchnorm \cite{ioffe2015batch} in both the generator and the discriminator, and removing fully connected layers for deeper architectures. 

After training DCGANs for image generation, the representations learned by DCGANs can be utilized for downstream tasks, by either fine tuning the discriminator features with an additional classification layer added on top and trained on $\mathcal{D}_l$, or by flattening and concatenating the learned features and training a linear classifier on top of them.

\subsubsection{SGAN}
DCGAN demonstrated the utility of the learned representations for SSL, but it has several undesirable properties. Using the learned representations of the discriminator after the fact doesn't allow for training the classifier and the generator simultaneously, doing this is more efficient, but more importantly, improving the discriminator improves the classifier, and improving the classifier improves the discriminator, which improves the generator. Semi-Supervised GAN (SGAN) \cite{odena2016semi} takes advantage of this feedback loop by allowing to learn a generative model and a classifier simultaneously, significantly improving the classification performance, the quality of the generated samples, and reducing training time.

Instead of a discriminator network outputting an estimated probability that the input image is drawn from the data distribution. For $C$ classes, SGAN consists of a discriminator with $C+1$ output, with per class output in addition to a \textit{fake} class output. Training an SGAN is similar to training a GAN; the only difference is using the labels to train the discriminator if the input $x$ is drawn for the labeled set $D_l$. The discriminator is trained to minimize the negative log-likelihood with respect to the given labels, and the generator is trained to maximize it.

\subsubsection{Feature Matching GAN}
Training GANs consists if finding a Nash equilibrium to a two-player non-cooperative game, with each player trying to minimize its cost function.
To solve this, GAN training consists of applying gradient descent on each player's cost simultaneously, but with such a training procedure, there is no guarantee of convergence.
Feature matching \cite{salimans2016improved} was proposed to encourage convergence. Feature matching addresses the instability of GANs by specifying a new objective for the generator that prevents it from over training on the current discriminator. Instead of directly maximizing the output of the discriminator, the new objective requires the generator to generate data that matches the first-order feature statistics between of the data distribution, \ie the hidden representations of the discriminator. For some activations $h(x)$ of a given intermediate layer, the new objective is defined as: 
\begin{equation}
\|\mathbb{E}_{x \sim p(x)} [h(x)]- \mathbb{E}_{z \sim p(z)} [h(G(z))] \|^{2}
\end{equation}

The problem of the generator mode collapse, where it always emits the same point, is still present even with feature matching because the discriminator processes each example independently, so there is no coordination between its gradients, and thus no mechanism to tell the outputs of the generator to become more dissimilar to each other. To avoid this, in addition to feature matching, a new technique called minibatch discrimination is also integrated into the training procedure to allow the discriminator to look at multiple data examples in combination, where the discriminator still classifies single examples as real or generated data, but it is now able to use the other examples in the minibatch as side information.

For SSL, similar to SGAN, the discriminator in feature matching GAN employs a $(C+1)$-class objective instead of binary classification, where true samples are classified into the first $C$ classes and generated samples are classified into the $(C+1)$-th fake class, the probability of $x$ being fake in this case is $p(y = C+1|G(z), D)$, corresponding to $1 - D(x)$ in the original GAN framework. The loss function for training the classifier then becomes $\mathcal{L}=\mathcal{L}_s + \mathcal{L}_u$ where:
\begin{equation}
\begin{aligned}
\mathcal{L}_s &=-\mathbb{E}_{x, y \sim p(x)_l} [\log p(y |x, y<K+1, D)] \\
\mathcal{L}_u &=-\mathbb{E}_{x \sim p(x)_u} \log [1- p(y=K+1|x, D)]-\mathbb{E}_{z \sim p(z)} \log [p(y=K+1|G(z), D))]
\end{aligned}
\end{equation}

The above objective is similar to the original GAN formulation by considering $p(y=K+1|G(z), D)$ to be the probability of fake samples, while the only difference is that the probability of true samples if split into $C$ sub-classes. This $(C+1)$-class discriminator objective lead to strong empirical results, and was later widely used to evaluate the effectiveness of generative models \cite{dumoulin2016adversarially,ulyanov2018takes}. The main drawback is that feature matching works well in classification but fails to generate indistinguishable samples, while the other objective of minibatch discrimination is good at realistic image generation but cannot predict labels accurately.

\subsubsection{Bad GAN}

Feature matching GAN formulation raises two questions. First, it is not clear why the formulation of the discriminator can improve the performance when combined with a generator. Second, it seems that good semi-supervised learning and a good generator cannot be obtained at the same time. Dai \etal \cite{dai2017good} addressed these questions by showing that for a $(C+1)$-class discriminator formulation of GAN-based SSL, good semi-supervised learning requires a \textit{bad} generator that does not match the true data distribution, but simply plays the role of a complement generator to help the discriminator obtain correct decision boundaries in high-density areas in the feature space.

To overcome the drawbacks of feature matching GANs, the new objective function of the generator is:
\begin{equation}
\min_{G}- \mathrm{H}(p_{G})+\mathbb{E}_{x \sim p_{G}} \log p(x) \mathbb{I}[p(x)>\epsilon]+\|\mathbb{E}_{x \sim p_{G}} h(x)-\mathbb{E}_{x \sim p(x)} h(x)\|^{2}
\end{equation}
where $p_{G}$ is the distribution induced by the generator $G$, $\mathbb{I}[\cdot]$ is an indicator function and $\epsilon$ is a threshold. The first term maximizes the entropy of generator to avoid the collapsing issues that are a clear sign of low entropy, but given that for implicit generative models, GANs only provide samples rather than an analytic density form, the entropy can either optimized in the input space \ie $\mathrm{H}(p_{G}(x))$ using variational inference or the feature space \ie $\mathrm{H}(p_{G}(h(x)))$ using a pull-away term (PT) \cite{zhao2016energy} as an auxiliary cost for the entropy. The second term enforces the generation of samples with low density in the input space by pushing the generated samples to move towards low-density regions defined by $p(x)$, this probability distribution over images is estimated using PixelCNN++ \cite{salimans2017pixelcnnpp} model, which pretrained on the training set, and fixed during semi-supervised training. The last term is the feature matching objective. This method substantially improves the performance of image classification over vanilla feature matching GANs on several benchmark datasets.

\subsubsection{Triple-GAN}
As discussed in Bad GAN, the generator and the discriminator (\ie the classifier) may not be optimal at the same time, since that for an optimal generator, \ie $p(x)=p_{g}(x)$, an optimal discriminator should identify $x$ as fake. Still, as a classifier, the discriminator should predict the correct class of $x$ confidently since $x \sim p(x)$, indicating that the discriminator and generator may not be optimal at the same time. Instead of learning a complement generator for classification, Triple-GAN \cite{chongxuan2017triple} is designed to achieve simultaneously a good generation of realistically-looking samples conditioned on class labels, and produce a good classifier with the smallest possible prediction error.

Triple-GAN consists of three components: (1) a classifier $C$ that characterizes the conditional distribution $p_c(y|x) \approx p(y|x)$; (2) a class-conditional generator $G$ that characterizes the conditional distribution in the other direction $p_g(x|y) \approx p(x|y)$; and (3) a discriminator $D$ that distinguishes whether a pair of data $(x,y)$ comes from the true distribution $p(x,y)$. All the components are parameterized as neural networks. The desired equilibrium is that the joint distributions defined by the classifier and the generator both converge to the true data distribution.

For $p(x)$ as the empirical distribution of inputs $x \in \mathcal{D}$ and $p(y)$ as a uniform distribution which is assumed to be the same as the distribution of labels on labeled data, the classifier produces pseudo-labels $p_c(y|x)$ given $x$, in this case, the examples $x$ and the pseudo-labels $y$ are drown from the joint distribution $p_c(x,y)=p(x)p_c(y|x)$.
Similarly, the generator produces examples $x = G(y, z)$, with $y \sim p(y)$ and the latent variables $z \sim p(z)$, the generated examples $x$ and labels $y$ are drown from the joint distribution $p_g(x,y) =p(y)p_g(x|y)$. These pseudo input-label pairs $(x,y)$ generated by both $C$ and $G$ are sent to the single discriminator $D$. The objective function is formulated as:
\begin{equation}
\begin{aligned}
\mathcal{L}=& \min _{C, G} \max _{D} E_{(x, y) \sim p(x, y)}[\log D(x, y)]+\alpha E_{(x, y) \sim p_{c}(x, y)}[\log (1-D(x, y))] \\
&+(1-\alpha) E_{(x, y) \sim p_{g}(x, y)}[\log (1-D(G(y, z), y))]
\end{aligned}
\end{equation}
where $\alpha \in [0, 1]$ is a constant that controls the relative importance of generation and classification. To properly leverage unlabeled data, an additional regularization is enforced on classifier $C$, consisting of minimizing the conditional entropy of $p_c(y|x)$, the cross-entropy between $p(y)$ and $p_c(y)$, and a consistency regularization with a dropout as the source of noise. In such a setting, the classifier achieves high accuracy with only very few labeled examples, while the generator produces state-of-the-art images, even when conditioned on $y$ labels.

Enhanced TripleGAN (EnhancedTGAN) \cite{wu2019enhancing} improves Triple-GAN by adopting a class-wise mean feature matching to regularize the generator and a semantic matching term to ensure the semantics consistency of the synthesized data between the generator and the classifier, further improving the state-of-the-art results in both SSL and instance synthesis.

\subsubsection{BiGAN}
One of the limitations of the traditional GAN framework is not being able to infer latent representations $z$ that can be used as rich representations of the data $x$ for a more efficient training. Unlike VAEs with an inference network (\ie decoder) $p(.)$ that can learn a variational posterior over latent variables, the generator is typically a directed, latent variable model with latent variables $z$ and observed variables $x$, making it unable to infer the latent feature representations for a given data point. BiGAN \cite{donahue2016adversarial} solves this by introducing an encoder $E$ as an additional component in the GAN framework, which maps data $x$ to latent representations $z$. The BiGAN discriminator $D$ discriminates not only in data space between $x$ and $G(z)$, but jointly in data and latent space, between pairs $(x,E(x))$ and $(G(z),z)$, where the latent component is either an encoder output $E(x)$ or a generator input $z$. A trained BiGAN encoder can then serve as feature extractor for downstream tasks. The BiGAN training objective is defined as a minimax objective:
\begin{equation}
\mathcal{L} = \min _{G, E} \max _{D} E_{x \sim p(x)}(\log D(x, E(x))) + E_{z \sim p(z)}(1 - \log D(G(z),z))
\end{equation}

Kumar \etal \cite{kumar2017semi} proposed Augmented-BiGAN, an improved version of BiGAN for SSL. The Augmented-BiGAN is similar to other GAN frameworks used for SSL, treating the generated samples as an additional class to the regular classes that the classifier aims to label, with an additional Jacobian-based regularization that is introduced to encourage the classifier to be robust to local variations in the tangent space of the input manifold. The BiGAN trained encoder is used in calculating these Jacobians, resulting in an efficient estimation of the tangents space at each training sample, and avoiding the expensive SVD-based method used in contractive autoencoders \cite{rifai2011manifold}.

%% file: graphbased.tex
\section{Graph-Based SSL}

Graphs are a powerful tool to model interactions and relations between different entities, in order to understand the represented system in both a global and local manner. In Graph-based SSL \cite{zhu2005semi} methods, each data point $x_i$, be it labeled or unlabeled, is represented as a node in the graph, and the edge connecting each pair of nodes reflects their similarity. Formerly, A graph $G(V, E)$ is a collection of $V=\{x_1, \ldots , x_n\}$ vertices or nodes and $E=\{e_{ij}\}_{i, j=1}^{n}$ edges. The $n \times n$ adjacency matrix $A$ of a graph $G$ describes the structure of the graph, with each element as a non-negative weight associated with each edge, if two nodes $x_i$ and $x_j$ are not connected to each other, then $A_{ij}$ = 0. 
The adjacency matrix $A$ can either be derived using a similarity measure between the data points \cite{zhu2003semi,iscen2019label}, or be explicitly derived from external data, such as a knowledge graph \cite{wijaya2013pidgin}, and provided as input.
Graph-based tasks can be broadly categorized into four categories \cite{goyal2018graph}: node classification, link prediction, clustering, and visualization. Graph methods can also be transductive or inductive in nature; transductive methods are only capable of producing labels assignments of the examples seen during training (\ie the unlabeled nodes of the graph), while inductive methods are more generalizable, and can be transferred and applied to unseen examples.
In this section, we will discuss node classification approaches, given that the objective in SSL is to assign labels to the unlabeled examples. Node classification approaches can be broadly grouped into methods which propagate the labels from labeled nodes to unlabeled nodes based on the assumption that nearby nodes tend to have the same labels \cite{azran2007rendezvous,zhu2003semi,zhou2004learning}, and methods which learn node embeddings based on the assumption that nearby nodes should have similar embeddings in vector space and then apply classifiers on the learned embeddings \cite{grover2016node2vec}. First, we start with some graph construction approaches and then discuss several popular methods for graph-based SSL.

\subsection{Graph Construction}
To apply graph-based SSL, we first need a graph. The graph can either be presented as an input in the form of an adjacency matrix $A$ or can be constructed to reflect the similarity of the nodes. A useful graph should reflect our prior knowledge about the domain and is the practitioner's responsibility to feed a good graph to graph-based SSL algorithms in order to produce valuable outputs (for more details, see Ch3 \& 7 \cite{zhu2003semi}).

In case we have limited domain knowledge about the dataset at hand, Zhu \etal \cite{zhu2003semi} describes some common ways to create graphs:
\begin{itemize}
\vspace{-0.5em}
\item \textbf{Fully connected graphs.} A simple form the graph can take is being fully connected with weighted edges between all pairs of data. With full connectivity, the derivatives of the graph w.r.t., the weights can be computed to update the weights of the edges, but the computational cost, in this case, will be high.
\item \textbf{Sparse graphs.} A sparse graph can be constructed so that each node is only connected to a few similar nodes, while the connections to dissimilar nodes are removed. Examples of sparse graphs are $k$NN graphs where nodes $i$ and $j$ are connected if $i$ is one of $k$-nearest neighbors \cite{von2007tutorial} of $j$ or vice versa.
A possible way to obtain the edge weight $A_{ij}$ between $x_i$ and $x_j$ is to use a Gaussian kernel \cite{bishop2006pattern}: $W_{i j}=\exp \{-\|x_i-x_j\|^{2} / 2 \sigma^{2}\}$
with a hyperparameter $\sigma$. Another approach is
$\epsilon$NN graphs where nodes $i$ and $j$ are connected if the distance $d(i, j) \leq \epsilon$. These graphs can be created using either the raw data or representations extracted from a trained network and updated iteratively (\eg CNN features \cite{iscen2019label}).
\end{itemize}

\subsection{Label Propagation}
The main assumption in label propagation is that the data points in the same manifold are very likely to share the same semantic label \cite{zhou2004learning}. To this end, label propagation \textit{propagates} labels of the labeled data points to the unlabeled data points according to the data manifold structures and the in-between node similarity.

In label propagation \cite{zhu2003semi,zhou2004learning,fujiwara2014efficient}, the labeling scores are defined as the optimal solution that minimizes the loss function. Let a $n \times C$ matrix $\hat{Y}$ corresponds to the new classification scores for each data point, where each row $\hat{Y}_i$ is a probability distribution over $C$ classes, and $Y$ is a $n \times C$ matrix containing the labels for the labeled data points, where each row $Y_i$ is a one-hot vector if $x_i$ is a labeled data point, and a vector of zeros otherwise. The loss function to be minimized for label propagation \cite{zhu2003semi} is:
\begin{equation}
\mathcal{L}=\frac{1}{2} \sum_{i, j = 1}^n A_{ij}(\hat{y}_i-\hat{y}_j)^{2}=\hat{Y}^{T} L \hat{Y}
\label{eq:zhu}
\end{equation}
where $L = D - A$ is the graph Laplacian matrix that measures the smoothness of the graph, with $D=\sum_{j=1}^{n} A_{ij}$ as the degree matrix, this loss function can be viewed as a graph Laplacian regularization which incurs a large penalty when similar nodes with a large weight $A_{ij}$ are predicted to have different labels $\hat{y}_i \neq \hat{y}_j$. By defining a $n \times n$ probabilistic transition matrix $P = D^{-1}A$, where $P_{ij}$ is the probability of transit from node $i$ to $j$, and spliting the matrices $P$, $Y$ and $\hat{Y}$ into labeled and unlabeled sub-matrices:
\begin{equation}
P=\left(\begin{array}{cc}
P_{l l} & P_{l u} \\
P_{u l} & P_{u u}
\end{array}\right) \quad Y=\left(\begin{array}{l}
Y_{l} \\
Y_{u}
\end{array}\right) \quad \hat{Y}=\left(\begin{array}{l}
\hat{Y}_{l} \\
\hat{Y}_{u}
\end{array}\right)
\end{equation}
the optimal solution for \cref{eq:zhu} is:
\begin{equation}
\begin{aligned}
\hat{Y}_{l} &=Y_{l} \\
\hat{Y}_{u} &=(I-P_{u u})^{-1} P_{ul} Y_{l}
\end{aligned}
\end{equation}
where $I$ is an identity matrix. The labeling score computation involves the matrix inversion operation, which is computationally heavy for large graphs. As an alternative, Zhu \etal \cite{zhu2003semi} propose an iterative approach to converge to the same solution:
\begin{enumerate}
\vspace{-0.5em}
\item Propagate $\hat{Y} \leftarrow P \hat{Y}$.
\item Preserve the labeled data $\hat{Y}_{l} = Y_{l}$.
\item Repeat from step 1 until convergence.
\end{enumerate}

Another similar label propagation algorithm was proposed by Zhou \etal \cite{zhou2004learning}, where, in addition to the contribution a node $i$ receives from its neighbors $j$, it receives an additional small contribution given by its initial value. In this case, the labels of the labeled nodes might change to better reflect the final labels, which can be helpful if the initial labels are noisy. The loss function in this instance is:
\begin{equation}
\mathcal{L}=\frac{1}{2} \sum_{i, j=1}^{n} A_{ij}\|\frac{\hat{Y}_i}{\sqrt{D_{ii}}}-\frac{\hat{Y}_i}{\sqrt{D_{jj}}}\|^{2}
+ (1 / \alpha - 1) \sum_{i=1}^{n}\|\hat{Y}_i-Y_i\|^{2}
\end{equation}
with a hyperparameter $\alpha$. The first and second terms in the loss function correspond to the smoothness constraint and the fitting constraint, respectively. The smoothness constraint results in labels that do not change too much between nearby points, while the fitting constraint forces the final labels of the labeled nodes to be similar to their initial value. The optimal solution that minimizes the loss function is:
\begin{equation}
\hat{Y} = (I - \alpha S)^{-1} Y
\end{equation}
where $S = D^{-1/2} A D^{-1/2}$. Similar the the first algorithm, Zhou \etal \cite{zhou2004learning} propose a less computationally expensive iterative approach:
\begin{enumerate}
\vspace{-0.5em}
\item Propagate $\hat{Y} \leftarrow \alpha S\hat{Y} + (1 - \alpha) Y$.
\item Repeat step 1 until convergence.
\end{enumerate}
It is worth noting that even though the iterative method is the standard approach for label propagation, it does not output the same labeling results as the the optimal solution.

\subsection{Graph Embedding}
The term graph embedding has been used in the literature in two ways: to represent an entire graph in vector space, or to represent each individual node in vector space \cite{goyal2018graph}. In this paper, we are interested in learning node embeddings since such a representation can be used for SSL tasks, such as node classification.
The goal of node embedding is to encode the nodes as low dimensional vectors that reflect their positions and the structure of their local neighborhood. These low dimensional embeddings can be viewed as encoding or projecting, nodes into a latent space, where geometric relations in this latent space correspond to interactions (\eg edges) in the original graph \cite{hamilton2017representation,hoff2002latent}. Factorization-based approaches such as Laplacian Eigenmaps \cite{belkin2002laplacian} and Locally Linear Embedding (LLE) \cite{cao2015grarep} are examples of algorithms based on this rationale, but they have scalability issues for large graphs, other more scalable embedding techniques which leverage the sparsity of real-world networks have been proposed. For example, LINE \cite{tang2015line} and HOPE \cite{ou2016asymmetric} attempt to preserve high order proximities (\eg the edge weights of a given node, and the similarity of edge weights of each pair of nodes). Intuitively, the goal of these methods is simply to learn embeddings for each node such that the inner product between the learned embedding vectors approximates some deterministic measure of graph proximity \cite{hamilton2017representation}.

Another family of method is random walks introduced by \cite{perozzi2014deepwalk} and its variants \cite{grover2016node2vec,ganguly2016author2vec,chen2018harp,abu2018watch}. Instead of using a deterministic measure of graph proximity like factorization-based approaches, these methods optimize the embeddings so that nodes have similar embeddings if they tend to co-occur within short random walks over the graph, making them especially useful when one can either only partially observe the graph or the graph is too large to measure in its entirety. Random walks consist of starting from a randomly samples node $x_0 \in \text{sample}(V)$, and then repeatedly sampling an edge to transition to the next node $x_{i+1} = \text{sample}(\mathcal{N}(x_{i}))$, with $\mathcal{N}(x_{i})$ as the neighboring nodes of $x_{i}$. The resulting sequence of random walks $x_{0} \rightarrow x_{1} \rightarrow x_{2} \rightarrow \ldots$ can then be passed to word2vec algorithm \cite{mikolov2013distributed} with the objective to embed each node $x_i$ within the random walk sequences to be close in the vector space to its neighboring nodes. With a context window of size $T$, with $T$ usually defined to be in the range $T \in \{2, \ldots, 10\}$, the representation of the anchor node $x_i$ is brought closer to the embeddings of its next neighbors $\{x_{i-T/2}, \ldots, x_i, \ldots, x_{i+T/2}\}$.

\begin{figure}
\centering
\includegraphics[width = 0.9\textwidth]{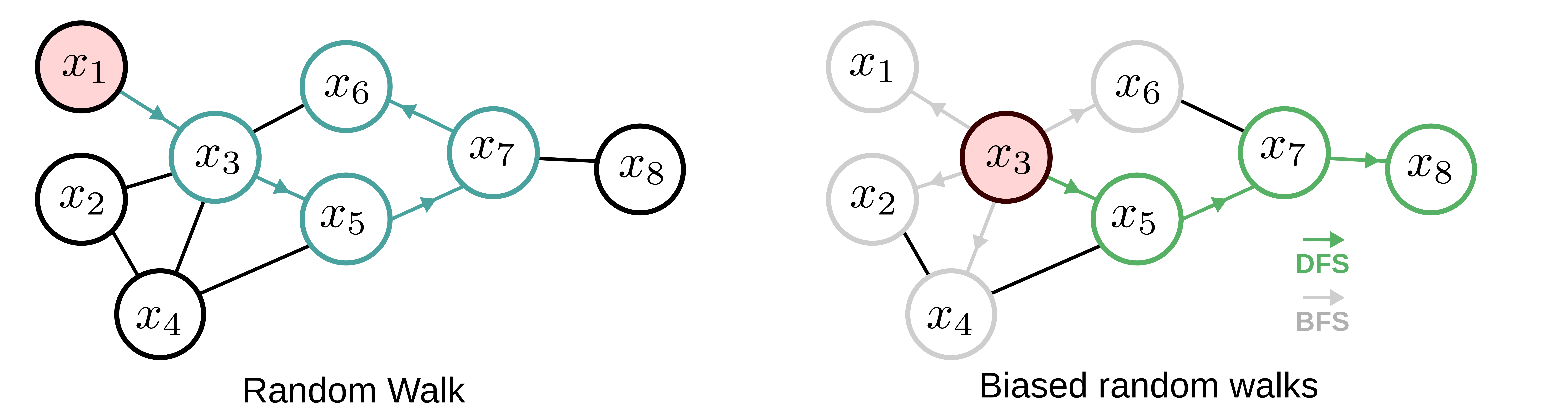}
\caption{\textbf{Random Walks.} \textit{Left.} An example of a random walk of length 4 starting
from node $x_1$: $x_1 \rightarrow x_3 \rightarrow x_5 \rightarrow x_7 \rightarrow x_6$
\textit{Right.} Breadth first search (BFS) and depth first search (DFS) strategies from node $x_3$.}
  \label{fig:randomwalks}
\end{figure}

Formally, random walk method learn embeddings $z_i$ of a given node $x_i$ so that:
\begin{equation}
p_T(x_j|x_i) \approx \frac{e^{z_i^{\top} z_j}}{\sum_{x_k \in V} e^{z_i^{\top} z_k}}
\end{equation}
where $p_T(x_j|x_i)$ is the probability of visiting $x_j$ on a length-$T$ random walk starting at $x_i$. To learn such embeddings, the following loss is optimized:
\begin{equation}
\mathcal{L}=\sum_{(x_{i}, x_{j}) \in \mathcal{RW}}-\log\left(\frac{e^{z_i^{\top} z_j}}{\sum_{x_k \in V} e^{z_i^{\top} z_k}}\right)
\label{eq:rw}
\end{equation}
where $\mathcal{RW}$ is the set of the length-$T$ generated random walks. Evaluating the loss is prohibitively expensive, since assessing the denominator requires a computation over all the nodes of the graph. Thus, different methods use different optimization to approximate the loss in \cref{eq:rw}. For example, DeepWalk \cite{perozzi2014deepwalk} uses a hierarchical softmax to compute the denominator, while node2vec approximates \cref{eq:rw} using negative sampling similar to word2vec. The different methods also differ in the construction of random walk, DeepWalk uses simple unbiased random walks over the graph, while node2vec introduces two random walk hyperparameters, $p$ and $q$, to smoothly interpolate between walks that are more akin to breadth-first or depth-first search (\cref{fig:randomwalks}). The hyperparameter $p$ controls the likelihood of the walk immediately revisiting a node, while $q$ controls the likelihood of the walk revisiting its neighborhood \cite{hamilton2017representation}.

For SSL, the learned embeddings can then be used as inputs to train a classifier over the labeled nodes and then applied over the unlabeled node. Alternatively, a cross-entropy term can be added to the unsupervised loss in \cref{eq:rw} for SSL based random walks, in order to jointly train a classifier on top of the node embeddings over the labeled nodes. For example, Planetoid \cite{yang2016revisiting} introduces a hyperparameter $r$ to control the sampled instances for a given training iteration, alternating between sampling random pairs from a given random walk for the unsupervised loss if $r < random$, and a couple of nearby labeled nodes with the same label for the supervised loss if $r \geq random$. The embeddings are trained using both the supervised and unsupervised loss, while the classifier is only trained with a supervised loss.

\subsection{Graph Neural Networks}
Random walks based method, with their expressivity (\ie incorporating both local and higher-order neighborhood information) and efficiency (\ie do not need to consider all node pairs when training), suffer from some limitations, such as the lack of parameter sharing where every node has its own unique embedding, and the inherent transductive nature of these approaches, in which the embeddings are only generated for nodes seen during training. This is especially problematic for evolving graphs, massive graphs that cannot be fully stored in memory, or domains that require generalizing to new graphs after training \cite{hamilton2017representation}. To solve these issues, a number of methods use deep neural networks based methods applied to graphs \cite{zhou2018graph,wu2020comprehensive,bacciu2019gentle}. DNGR \cite{cao2016deep} and SDNE \cite{wang2016structural} propose the first application of deep networks for graphs by using deep autoencoders \cite{hinton2006reducing} in order to compress the information about a node's local neighborhood. A high dimensional representation $s_i \in \mathbb{R}^{|V|}$ of a node $x_i$, which describes the proximity of node $x_i$ to all other nodes in the graph is first extracted, and then fed through an autoencoder for dimensionality reduction and trained using a reconstruction loss. After training, the bottleneck low dimensional representation is then used as an embedding for $x_i$. However, these approaches suffer from similar limitations as random walks methods, with inputs of size $|V|$, which can be extremely costly and even intractable for large graphs, in addition to their transductive nature.

\begin{figure}
\centering
\includegraphics[width = 0.7\textwidth]{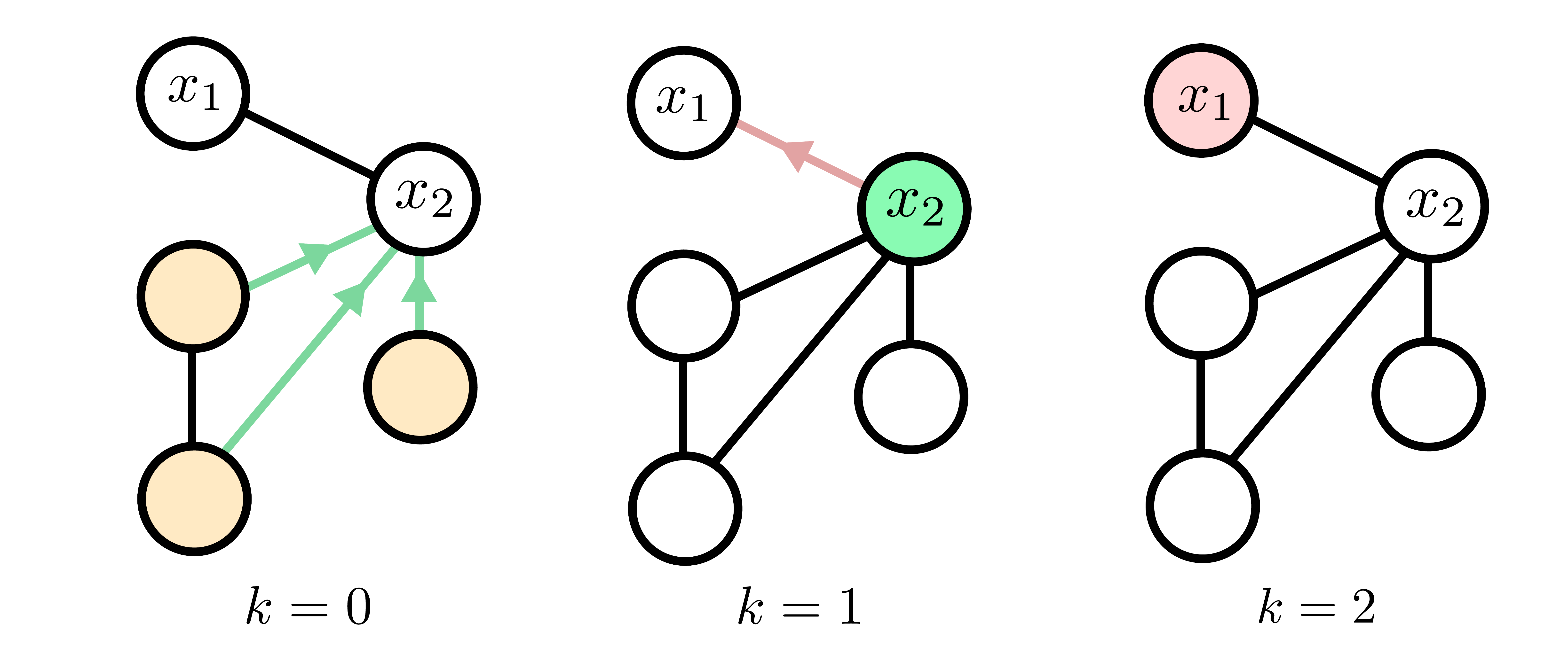}
\caption{\textbf{Context Aggregation.} An example of a three-step context aggregation. The context of $x_1$ at $k=2$ depends not only
on its neighboring node $x_2$, but also the neighbors of $x_2$ due to the first aggregation step.}
\label{fig:aggregation}
\end{figure}

Several recent node embedding approaches aim to solve the main limitations of the random walks and autoencoder based methods by designing functions that rely on a node's local neighborhood (\cref{fig:aggregation}), but not necessarily the entire graph. Unlike the previously discussed methods, graph neural networks use the node features, \eg profile information for a social network or even simple statistics such as node degree \cite{hamilton2017inductive} or one-hot vectors \cite{kipf2016variational}, to generate the embeddings. These methods are often called convolutional because they represent a node as a function of its surrounding neighborhood, similar to CNNs \cite{hamilton2017representation}. The training procedure starts by initializing the first hidden states $h_i^{0}$ using the nodes features $x_i$: $h_i^{0} \leftarrow x_i, \forall x_i \in V$. For $K$ training iterations, at each step, the hidden stated are updated by aggregating the hidden states of the neighboring nodes, with an $\operatorname{AGGREGATE}$, $\operatorname{COMBINE}$, a non-linearity $\sigma$, and a $\operatorname{NORMALIZE}$ functions as follows \cite{hamilton2017inductive}:
\begin{itemize}\vspace{-0.5em}
\item[] For $k=1 \ldots K$:
	\begin{itemize}\vspace{-0.5em}
	\item[] For $x_i \in V$:
		\begin{enumerate}
			\item $h^{\prime} \leftarrow \operatorname{AGGREGATE}_{k}(\{h_{j}^{k-1}, \forall x_j \in \mathcal{N}(x_i)\})$
			\item $h_{i}^{k} \leftarrow \sigma(W^{k} \cdot \operatorname{COMBINE}(h_{i}^{k-1}, h^{\prime}))$
			\item $h_{i}^{k} \leftarrow \operatorname{NORMALIZE}(h_{i}^{k} )$
		\end{enumerate}
	\end{itemize}
\end{itemize}
and at the end, the embeddings $z_i$ of node $x_i$ are the final hidden states: $z_i \leftarrow h_i^{K}$.
The aggregation function and the set of trainable parameters $W^{k}, \forall k \in[1, K]$ specify how to aggregate the local neighborhood information. The different approaches such as GCN \cite{scarselli2008graph,tang2015line,kearnes2016molecular,kipf2016semi}, GraphSAGE \cite{hamilton2017inductive} and GAT \cite{velivckovic2017graph} follow the same procedure but differ primarily in how the aggregation, the combination and the normalization are performed. For example, GraphSAGE uses concatenation as a combination function and experiment with various general aggregation functions, \ie the element-wise mean, max-pooling, and LSTMs, while GCN uses a weighted sum as a combination function and element-wise mean as an aggregate. The weight are then trained using an unsupervised loss similar to random walks based methods, and for SSL \cite{kipf2016semi}, a classifier is trained on top of the node embeddings (\ie the final hidden state) to predict the class labels for the labeled nodes, which can then applied on the unlabeled nodes for node classification.

%% file: unsupervised.tex
\section{Self-Supervision for SSL}

Self-supervised learning \cite{weng2019selfsup,jing2019self} is a form of unsupervised learning, where the model is trained using a standard supervised loss, but on a \textit{pretext task} where the supervision comes from the data itself. The objective, in this case, is not to maximize final performance on the pretext task, but rather to learn rich and transferable features for downstream tasks. 
A variety of pretext tasks were proposed, where the model is first trained on one or multiple tasks with unlabeled examples, the resulting model is either used for generating representations for the raw data, which are utilized for training a shallow classifier on $\mathcal{D}_l$, or directly fine-tuned for a downstream task with labeled images. Examples of such pretext tasks for computer vision are:
\begin{itemize}\vspace{-0.5em}
\item \textbf{Exemplar-CNN} \cite{dosovitskiy2015discriminative}. For a given image, a set of $N$ patches are generated using different transformations, all these patches are then considered as a separate class, and the model is trained to predict the correct class for a given input patch.
\item \textbf{Rotation} \cite{gidaris2018unsupervised}. A given rotation out of four possible rotations of multiple of $90^{\circ}$, \ie $[0^{\circ}, 90^{\circ}, 180^{\circ}, 270^{\circ}]$, is applied to the input image, and the model is trained to predict the correct rotation that was applied.
\item \textbf{Patches} \cite{doersch2015unsupervised}. A first patch is randomly extracted from the input image, this patch is considered as the center, and eight different neighboring and non-overlapping patches are extracted with small jitters at the eight neighboring locations, the model is then trained to predict the position of one of the second patches with regard to the first one. Other versions of this pretext task were proposed, such as jigsaw puzzle \cite{noroozi2016unsupervised} where a random permutation of the nine patches are fed into the model, and the objective is to predict the correct permutation that was applied to get the correct ordering of the patches.
\item \textbf{Colorization} \cite{zhang2016colorful}. The input image is first transformed from RGB to Lab color space, an input with only the luminance information contained within the L component or the color information with ab components is fed into the model, and the objective is to predict the rest of the information, \ie either the luminance or the coloring of the image. The task can either be considered as a regression problem or a classification problem by quantizing the Lab color space.
\item \textbf{Contrastive Predictive Coding} \cite{oord2018representation}. Using a contrastive loss based on Noise Contrastive Estimation \cite{gutmann2010noise} and its recent versions such as Momentum Contrast \cite{he2019momentum} and SimCLR \cite{chen2020simple}, the model is trained to differentiate between positive and negative samples, the positives can be a given input image and its transformed versions, or a given patch and its neighboring patches, while the negatives are randomly sampled images or patches. 
\end{itemize}

Such pretext tasks can easily be utilized for SSL, where the model is trained on the whole dataset on the pretext task with self-supervision, and then adapted to the labeled set $\mathcal{D}_l$ using the standard cross-entropy loss, either simultaneously as demonstrated by \cite{zhai2019s4l,berthelot2019remixmatch} with rotation as the pretext task, or iteratively, by first training the model using self-supervision and then fine tuning it on $\mathcal{D}_l$ as demonstrated by \cite{chen2020simple,chen2020big} using contrastive learning.